\documentclass[msom]{informs4} 
\usepackage{amsmath}
\DoubleSpacedXI
\usepackage{natbib}
\usepackage{color}
\usepackage{subfig}
\usepackage{algorithm} 
\usepackage{algorithmic} 
\usepackage{multirow} 
\usepackage{booktabs}
\usepackage{epstopdf} 
\bibpunct[, ]{(}{)}{,}{a}{}{,}%
\def\bibfont{\small}%
\def\bibsep{\smallskipamount}%
\def\bibhang{24pt}%

\makeatletter
\renewcommand*\env@matrix[1][c]{\hskip -\arraycolsep
	\let\@ifnextchar\new@ifnextchar
	\array{*\c@MaxMatrixCols #1}}
\makeatother

\TheoremsNumberedThrough     
\ECRepeatTheorems

\EquationsNumberedThrough    

\MANUSCRIPTNO{}
\newcommand{\hl}[1]{\textcolor{black}{#1}}
\begin{document}


\RUNAUTHOR{Liu, Shen, and Ji}

\RUNTITLE{Bike Lane Planning}

\TITLE{Urban Bike Lane Planning with Bike Trajectories: Models, Algorithms, and a Real-World Case Study}

\ARTICLEAUTHORS{%
	\AUTHOR{Sheng Liu}
	\AFF{Rotman School of Management, University of Toronto, \EMAIL{sheng.liu@rotman.utoronto.ca}}
	\AUTHOR{Zuo-Jun Max Shen}
	\AFF{Department of Industrial Engineering and Operations Research and Department of Civil and Environmental Engineering, University of California, Berkeley, \EMAIL{maxshen@berkeley.edu}}
	\AUTHOR{Xiang Ji}
	\AFF{Department of Electrical Engineering and Computer Sciences, University of California, Berkeley, \EMAIL{xji2020@berkeley.edu }}
} 

\ABSTRACT{We study an urban bike lane planning problem based on the fine-grained bike trajectory data, \hl{which is made available by smart city infrastructure such as bike-sharing systems}. The key decision is where to build bike lanes in the existing road network. As bike-sharing systems become widespread in the metropolitan areas over the world,  bike lanes are being planned and constructed by many municipal governments to promote cycling and protect cyclists. Traditional bike lane planning approaches often rely on surveys and heuristics. We develop a general and novel optimization framework to guide the bike lane planning from bike trajectories. We formalize the bike lane planning problem in view of the cyclists' utility functions and derive an integer optimization model to maximize the utility. \hl{To capture cyclists' route choices, we develop a bilevel program based on the Multinomial Logit model. } We derive structural properties about the base model and prove that the Lagrangian dual of the bike lane planning model is polynomial-time solvable. \hl{Furthermore, we reformulate the route choice based planning model as a mixed integer linear program using a linear approximation scheme.}  We  develop tractable formulations and efficient algorithms to solve the large-scale optimization problem. Via a real-world case study with a city government, we demonstrate the efficiency of the proposed algorithms and quantify the trade-off between the coverage of bike trips and continuity of bike lanes. We show how the network topology evolves according to the utility functions and highlight the importance of \hl{understanding  cyclists' route choices}. The proposed framework drives the data-driven urban planning scheme in smart city operations management.}

\KEYWORDS{Smart city, Urban planning, Facility location, Optimization, Integer programming}

\maketitle

\section{Introduction}\label{sec:intro}
Urbanization is a global trend. More than half of the world's population now lives in towns and cities. It is projected that about 56\% of the developing world and 82\% of the developed world will be urbanized by 2030, which equals to 5 billion people \citep{Econ2012}. Much of this urbanization will unfold in Africa and Asia, bringing huge social, economic and environmental transformations \citep{UNPF2018}. The main challenges brought by the fast urbanization include traffic congestion and air pollution. The growing number of cars exceeds the city's traffic carrying capacity and thus causes severe traffic congestion around the city. In the meanwhile, the vast amount of emissions generated by moving cars worsens the air quality and poses serious public health problems. For Beijing, cars accounted for 30\% of city's self-generated pollutants contributing to air pollution \citep{SCMP2018}.

In addition to driving cars, cycling is a popular urban transit mode for daily commute. For instance, 30\% of people go to work by riding bikes in Cambridge, UK \citep{UK2015}. According to \cite{US2012}, 864,883 people cycled to work in the United States. Cycling is promoted by many countries as it benefits  city residents in many different aspects. First, cycling is free of pollution and the large-scale adoption of cycling can benefit the urban environment. Second, the popularity of cycling could alleviate the traffic congestion and improves the overall traffic condition. Third, cycling is also a more affordable and healthy transit mode than driving cars. As shown by the \cite{Copen2010}, mortality is reduced by 30\% in adults who cycle to and from their workplace on a daily basis, which can translate to huge savings of health care cost. \hl{During the COVID-19 pandemic, bike ridership has increased by 21\% in US urban areas as people are shifted from public transit to this open-air mobility mode \citep{latimes2020}.}  

To promote cycling and reduce bike crashes, bike lanes have been planned and constructed by city planners. In Amsterdam, for example, there are 250 miles of dedicated bike lanes in use. In the U.S., city governments are rolling out better bike lanes in a large scale. For instance, Chicago has planned to build 50 miles of bike lanes by 2019 \citep{Chi2015}. Cities from developing countries, such as China and India, are also investing heavily in protected bike lanes \citep{Sea2016}. \hl{To accommodate the booming cycling demand during the 2020 pandemic, municipal governments across Canada such as Toronto Council have fast-tracked their bike lane expansion plans \citep{cbc2020}.} The construction of bike lanes improves the safety for cyclists, car drivers as well as pedestrians. It is reported that fatality rate is less than a tenth as high in the countries with well designed cycling road networks as in the countries without cycling friendly infrastructures \citep{pucker2001cycling}. 

Traditionally, bike lanes are planned based on experience and surveys. As smart phones are widely used, the GPS human mobility data becomes more available and makes it possible to utilize this fine-grained data in bike lane planning. \hl{Most recently, smart city infrastructure such as station-based and dock-less bike sharing systems are expanding quickly across the world. Some examples include Citi Bike (New York, USA), Santander Cycles (London, UK) for station-based systems and Lime Bike (USA), Mobike (China) for dock-less systems. Bay Wheels, the bike sharing system operated in San Francisco Bay Area, has maintained a mixed fleet of station-based and dock-less bikes (mainly electric models). In these systems, people can use smart phones to pick up and drop off bikes at docking stations or arbitrary locations with a built-in smart lock. Bikes in the dock-less system are often embedded with GPS devices (thus often called "smart bikes", e.g., see \citealt{hu2017no}) so the spatial bike trajectories can be tracked. These bike trajectories record the detailed travel pattern of bike riders, which can  then be used to understand the route choices made by the cyclists. Compared to the survey statistics and conventional origin-destination data, the trajectory data generated from the smart city infrastructure reveals cyclists' footprints on the road network, which are otherwise difficult to collect and unknown to policy makers. The trajectory data is essential to bike lane planning as most bike lanes are constructed along the existing road network. }

\hl{The bike trajectory data is becoming increasingly accessible to city planning agencies and researchers as bike sharing system operators are pursuing open data policies. For instance, Mobike is fully collaborating with municipal governments to facilitate urban management \citep{hangzhou2017, mobike2018}. The Los Angeles Department of Transportation has created an open-source software platform called Mobility Data Specification (MDS) that allows authorities to collect data directly from mobility service providers (including dock-less bikes and scooters) in real time\footnote{The detailed description about MDS can be found at \url{https://ladot.io/wp-content/uploads/2018/12/What-is-MDS-Cities.pdf}}. MDS is now used by more than 50 American cities and dozens more around the globe \citep{mds2020}. How to analyze and utilize this high-volume trajectory information to accomplish the smart city vision will be a critical challenge to city planners.}

This paper presents and solves the bike lane planning problem using the detailed bike trajectory data. The bike lane planning problem decides on which road segments of the existing road network to construct bike lanes, aiming to balance two main objectives: 1) \textit{coverage}: cover as many bike trips as possible and 2) \textit{continuity}: build more continuous bike lanes to minimize interruptions. We summarize our main results and contributions as follows.

\begin{enumerate}
	\item We present the bike lane planning model in view of the cyclist's utility functions based on the trajectory data. We start from a simple adjacency-continuity utility function and then discuss the general class of utility functions. The choice of the utility functions is flexible to characterize the trade-off between the coverage and continuity objectives. To the best of our knowledge, our work is the first to formalize the general bike lane planning model built upon the bike trajectory data. 
	\item For the simple adjacency-continuity utility function, we show that the resulting bike lane planning model has a supermodular objective function and admits an efficient mixed integer linear program (MILP) formulation. For the general utility functions, we show that under reasonable conditions, the objective function is also supermodular and the resulting problem yields a polynomial-time solvable Lagrangian dual. Furthermore, we provide a linear programming approach to the Lagrangian relaxation subproblem and propose an efficient algorithm for the general utility functions. 
	\item \hl{To capture the interaction between cyclists' route choices and bike lane design, we propose a route choice based bike lane planning model and formulate it as a bilevel program. By exploiting the structure of the lower-level problem, we reformulate the bilevel program as a single-level MILP, which is asymptotically exact and computationally tractable.}
	\item We present a real-word case study based on the collaboration with an urban planning institution and a dock-less bike sharing company. We collect and preprocess a large bike trajectory data set, and test our models and algorithms via extensive numerical experiments. The numerical experiments validate the efficiency of the proposed algorithms and deliver insightful comparison results between different models. We show how the topology of bike lane network would change according to the utility functions and provide quantitative measures to analyze the trade-off between coverage and continuity. We also highlight the importance of understanding cyclists' route choice behaviors in the bike lane planning. 
\end{enumerate}

The remainder of the paper is organized as follows. Section \ref{sec:work} reviews the related literature. \hl{Section \ref{sec:mod} develops the bike lane planning model and presents tractable formulations and structural results, which motivates efficient algorithms. Section \ref{sec:case} is devoted to the real-world case study and numerical experiments that deliver managerial insights for policy makers.} Section \ref{sec:con} concludes the paper and presents several future research directions.

\section{Related Literature}\label{sec:work}
Smart city operations have received growing attention from the operations management community. Different aspects of the smart city movement have been studied, including smart grid, smart transportation, and smart retail. The ``smart" parts of these aspects lie in the innovative technologies that can disrupt the present urban environment and city operations, or new data collection and data analytics tools to inform better understanding and decisions for city management. For instance, \cite{zhang2020values} study a promising scheme where vehicle-to-grid (V2G) electricity selling is integrated in electric vehicle sharing systems, as enabled by new technological development. They provide important strategic planning and operational tools to support the advancement of this scheme. Shared mobility, where shared passenger cars are deployed to provide ride and logistics services, has been examined from various perspectives such as pricing (\citealt{taylor2018demand, bai2018coordinating, guda2019your}), admission control (\citealt{afeche2018ride}), repositioning \citep{he2020robust} and last-mile delivery \citep{qi2018shared}, among others. In retailing, recent development of IT and big data tools have enabled innovative omni-channel strategies \citep{gao2016omnichannel, harsha2019dynamic} and data-driven pricing and logistics policies \citep{perakis2018joint, glaeser2019optimal, liu2019data}. We refer readers to \cite{mak2018enabling} and \cite{qi2019smart} for thorough reviews of papers in smart city operations.  Our paper is among the first to develop rigorous analytical models to tackle an urgent city planning problem built upon new mobility data sources, and thus promotes the smart city vision.

There has been an increasing focus on the planning and control of bike facilities, in particular, the optimization of bike-sharing systems in recent years. 
The majority of the existing literature tackles the operational problems of bike sharing systems, such as balancing bike stock levels among bike stations to satisfy temporal and spatial demand. For example, \cite{shu2013models} address the bike deployment and redistribution problems with network flow formulations. 
\cite{o2015data} study similar rebalancing problems as well as routing and clustering problems with different integer programming techniques.  In addition, empirical approaches have been applied to shed lights on the optimal configuration of bike share systems \citep{kabra2019bike}. 
However, few papers consider the roads used by cyclists to travel between bike stations, which is another fundamental deciding factor of a bike-sharing system's success. Noticeably, the above three papers only utilize the origin-destination information of bike rides while our paper incorporates a fine-grained bike trajectory data set, which makes it possible to capture more detailed travel patterns of bike riders. 

Traditionally, bike lane planning is not built upon reliable real-world data. A prominent approach of traditional bike-lane planning is to select road segments (on which to construct bike lanes) by evaluating them with respect to a set of predetermined guidelines and criteria. \cite{dondi2011bike} propose a context sensitive approach and make the selection based on the analysis of the visual effects, environmental contexts, and traffic considerations of road segments. \cite{rybarczyk2010bicycle} select road segments with a modified simple additive weighting method to calculate an overall score for each road segment based on its rankings among all road segments for each criterion. 
Since the guidelines and criteria in these papers are decided based on expertise or specific needs, there is a potential risk of subjective bias. Another classical approach is to select road segments based on cyclists survey results that reveal their preferences. One such approach is built on stated preference surveys, in which respondents are asked to imagine their preferences in some hypothetical cases \citep{tilahun2007trails,stinson2003commuter,hunt2007influences}. However, results from the stated preference surveys could be inaccurate as they do not represent cyclists' actual choices. Researchers also adopted revealed preference surveys, in which respondents reveal their actual choices \citep{sener2009analysis,howard2001cycling,hyodo2000modeling}. Results from revealed preference surveys are also subject to sampling biases and small sample biases. 

In addition to subjective and survey based methods, there are a few papers that develop analytical methods for the bike lane planning problem. 
\cite{lin2011strategic} consider a strategic designing problem for bike sharing, in which the location of bike stations and bike lanes are decided to minimize the system-wide operating cost. Their model only accounts for the origins and destinations while ignoring the road network. As a result, their model can not be directly used to guide the practical bike lane construction. Most recently, \cite{bao2017planning} propose several heuristics to decide the bike lane locations by maximizing a specific score function under budget and connectivity constraints. 
Our paper, on the other hand, proposes a general and tractable modeling framework for the bike lane planning, and derives structural results about the resulting planning problem. The structural results motivate efficient algorithms with empirically validated performance. 

From a modeling perspective, our paper is related to the general class of facility location problems. While many traditional facility location models are focused on locating warehouses/distribution centers/retailing stores to maximize profits or minimize operational costs \citep{snyder2019fundamentals}, we consider locating bike lanes on the existing road network to maximize riders' utility. Hence our paper echoes the emerging location models in nonprofitable operations and healthcare operations \citep{cho2014simultaneous, chan2017robust}.

\section{Bike Lane Planning Model}\label{sec:mod}
We first introduce the bike lane planning model based on bike trajectory data with a specialized utility function in Subsection \ref{sec:ac}. We then analyze the structural properties of the model and discuss a class of general utility functions in Subsection \ref{sec:gu}. \hl{Lastly, we extend the model to consider cyclists' route choices in Subsection \ref{subsec:behav}.}
\subsection{Adjacency-Continuity Utility Maximization} \label{sec:ac}
Let $V$ be the set of all road segments that have been visited by bike riders in our data set and $M$ be the set of cyclists. We say two road segments $i$ and $j$ are neighbors, $(i,j)\in N$, if they are connected. We can also interpret $(i,j)\in N$ as road intersections. For each road segment $i\in V$, we use $d_i$ to denote the associated number of bike trips. Similarly, each pair of connected road segment $(i,j) \in N$ is also associated with the number of trips that go through the corresponding intersection, $d_{ij}$.

We consider two main objectives of designing bike lanes inspired by the literature and our communication with the biking community:
\begin{enumerate}
	\item Constructed bike lanes should be able to cover as many bike trips as possible. (\textit{coverage objective)}
	\item Constructed bike lane network should enable continuous and smooth riding experience for cyclists. (\textit{continuity objective})    
\end{enumerate} 
Continuity is preferred for both cyclists and urban planning institutions. \cite{krizek2005end} show that the discontinuity of bike lanes can cause great discomfort to cyclists. The discontinuity at intersections also generates potentially higher crash risks. For the government, discontinuous bike lanes can pose management challenges as well as construction difficulties. The coverage objective and the continuity objective are often conflicting with each other, as shown in \cite{bao2017planning}. Maximizing only the coverage may lead to very dispersed bike lanes while maximizing only the continuity can leave many cyclists uncovered. So we need to find the ideal trade-off between the two objectives.

Now we formalize the two objectives from the cyclist's perspective. For a cyclist $m$, we use $r_m = \{i_m^1, \dots, i_m^{n_m}\}$ to denote the ordered set of road segments (i.e. trajectory) she traveled through, where $i_m^1, \dots, i_m^{n_m}\in V$ and $n_m$ is the number of road segments traveled by $m$ ($|r_m|$). The cyclist receives a positive utility if there is a bike lane on the road segment, i.e. $x_i=1$. Also, she gets an additional $\lambda$ utility if the bike lanes are continuous along an intersection, i.e. $x_i= x_{i+1}$. So her gained utility of traveling through $r_m$ from the bike lane construction plan $x$ is 
\[v_x(r_m) = \sum_{k=1}^{n_m} x_{i_m^k} + \lambda \sum_{k=1}^{n_m-1} x_{i_m^k} x_{i_m^{k+1}}.\]
Summing over the utility functions of all cyclists gives
\begin{align*}
\sum_{m\in M} v_x(r_m) = \sum_{i\in V} d_ix_i + \lambda \sum_{(i,j)\in N} d_{ij} x_ix_j
\end{align*}
where $d_i$ is the number of trajectories going through road segment $i$ and $d_{ij}$ is the number of trajectories going through the intersection $(i,j)$. So $\sum_{i\in V} d_ix_i$ stands for the total number of covered road segments (with bike lanes) weighted by the travel demand, and $\sum_{(i,j)\in N} d_{ij} x_ix_j$ is the additional continuity utility for two adjacent bike lanes weighted by the travel demand. Since the above utility function measures the continuity utility along two adjacent road segments, we call this utility function as the adjacency-continuity (AC) utility function. Note that in our discussion we treat each road segment equally regardless of the length for the ease of exposition and the analysis extends straightforwardly to consider the impact of length on the utility.

Based on the AC utility function, we propose a bike lane planning model that takes into account both the coverage and the continuity requirement. Let $x_i\in\{0,1\}$ denote the bike lane construction decision variable: $x_i=1$ if a bike lane is planned at road segment $i$ and $x_i=0$ otherwise. And we use $c_i$ to denote the construction cost of building a bike lane on road segment $i$. The bike lane planning model (BL) can be formulated as an integer program (IP):
\begin{align}
\tag{BL-AC}
\max_{x}\quad &  \sum_{i\in V} d_ix_i + \lambda \sum_{(i,j)\in N} d_{ij} x_ix_j, \label{obj:form1}\\
\mbox{s.t.} \quad & \sum_{e_i\in V} c_ix_i \leq B, \label{budget:form1}\\
& x_{i}\in\{0,1\},\quad \forall i\in V.
\end{align}
The objective function measures the cyclists' welfare from bike lanes. The parameter $\lambda\geq0$ determines the relative continuity benefit to the cyclist. A higher $\lambda$ means continuity is more desirable and thus a more continuous bike lane network would be proposed. Constraint (\ref{budget:form1}) is the budget constraint that ensures the total construction cost does not exceed the allowable government budget $B$.  
In practice, there may be other constraints that limit the construction of bike lanes in certain regions, which can be incorporated as needed. Furthermore, it is possible that building bike lanes along certain roads may reduce the capacity for car traffic flows, and hence worsening the traffic condition. To account for this, we can add additional penalty terms to our objective function. The detailed modeling of this traffic impact is out of the scope of this paper and we leave it for future research.

\hl{The above model is set up for the scenario when a city is planning bike lanes from scratch. For cities that already have bike lanes in use (such as New York City), the proposed model can help guide the network expansion: we can fix the decision variables $x_i=1$ and set $c_i=0$ for road segments that already have bike lanes in use, and $B$ is equal to the expansion budget. The obtained solution will then identify road segments to build new bike lanes.  }

\subsubsection{Analysis} When $\lambda=0$, the problem reduces to the classical Knapsack problem, which is NP-hard.  Otherwise, the problem is a special case of the 0-1 quadratic knapsack problem (QKP), where the coefficient matrix for the quadratic terms have a sparse structure. If $d_{ij}$ is separable and can be decomposed as $d_{ij} = \tilde{d}_i\tilde{d}_j$, then the objective function is referred to as the \textit{half-product} function. The maximization of the half-product function over a knapsack polytope is known to admit a \textit{Fully Polynomial-Time Approximation Scheme} (FPTAS). However, no FPTAS is available for the general non-separable objective function. 

Nevertheless, similar to the general QKP, it can be shown that the objective function with $d_{ij}\geq 0$ is supermodular.
\begin{lemma}\label{lem:1}
	When $\lambda\geq0$ and $d_{ij}\geq0$ for all $(i,j)\in N$, the objective function of \ref{obj:form1} is supermodular.
\end{lemma}

The proofs of Lemma \ref{lem:1} and other results in this section are presented in the Appendix. The supermodularity result implies that the problem can be solved efficiently without the budget constraint. So we can adopt the Lagrangian relaxation methodology to relax the budget constraint. The resulting Lagrangian dual is given as
\begin{equation}
\min_{u\geq 0} \Phi(u),
\end{equation}
where
\begin{align}
\Phi(u) = \max_{x_i\in\{0,1\}}\quad&  \sum_{i\in V} d_ix_i + \lambda \sum_{(i,j)\in N} d_{ij} x_ix_j - u(\sum_{i\in V} c_ix_i - B). 
\end{align}
Here $\Phi(u)$ is the Lagrangian relaxation of \ref{obj:form1} for \hl{dual variable} $u\geq0$. Based on the result of \cite{gallo1989supermodular} and \cite{chaillou1989best}, $\Phi(u)$ can be solved in polynomial time.
\begin{proposition}\label{prop:1}
	When $\lambda\geq0$ and $d_{ij}\geq0$ for all $(i,j)\in N$, the Lagrangian dual of \ref{obj:form1}  can be solved in polynomial time.
\end{proposition}

More specifically, one can show that the Lagrangian dual is a piece-wise linear convex function with at most $|V|$ break points. And each $\Phi(u)$ is equivalent to a maximum flow problem \citep{chaillou1989best}. There are other Lagrangian relaxation methods for QKP that rely on relaxing different sets of constraints (e.g. \citealt{caprara1999exact}), we refer readers to \cite{pisinger2007quadratic} for an extensive review.  Since the Lagrangian dual only provides an upper bound, we may still need to perform branch and bound to get an exact solution.  Alternatively, we can get an equivalent mixed integer linear programming (MILP) formulation to \ref{obj:form1}, which can deliver satisfactory computational performance with the commercial MILP solvers.

\subsubsection{MILP Formulation}
Although the objective function of \ref{obj:form1} is nonlinear, we can linearize the product terms in \ref{obj:form1} by replacing $x_ix_j$ with $y_{ij}$, and derive the following MILP formulation for bike lane planning:
\begin{align}
\tag{BL-AC-MILP}
\max_{x}\quad &  \sum_{i\in V} d_ix_i + \lambda \sum_{(i,j)\in N} d_{ij}  y_{ij}, \label{obj:form2}\\
\mbox{s.t.} \quad & y_{ij} \geq x_i + x_j - 1,\quad \forall (i,j)\in N, \label{const:ilp1}\\
&y_{ij}\leq x_i,\quad \forall (i,j)\in N,\label{const:ilp2}\\
&y_{ij}\leq x_j,\quad \forall (i,j)\in N,\label{const:ilp3}\\
\quad & \sum_{i\in V} c_ix_i \leq B, \label{budget:form2}\\
& 0\leq y_{ij}\leq 1,\quad\forall (i,j)\in N,\\
& x_{i}\in\{0,1\},\quad \forall i\in V.
\end{align}
Constraints (\ref{const:ilp1})-(\ref{const:ilp3}) ensure that $y_{ij}=1$ if $x_i=x_j=1$ and $0$ otherwise. Constraints (\ref{const:ilp1}) are redundant when $d_{ij}$'s are positive. \ref{obj:form2} is ready to be solved using commercial solvers such as Gurobi and CPLEX. 

\subsection{General Utility Functions} \label{sec:gu}
The AC utility function assumes the continuity utility only applies to two adjacent road segments. However, to the cyclists, their utility may depend on the size (length) of continuous bike lanes, which can not be captured by the AC function. It is often the case that a longer continuous bike lane is more preferable than a few shorter continuous bike lanes. On the other hand, the developed bike lane system from maximizing over the AC utility may consist of many relatively short bike lanes since only pair-wise continuity is rewarded.  
The following example illustrates the case where the AC utility may fail to reflect the actual utility from cyclists.

\begin{example}
Consider a cyclist riding through $r=\{1,2,3,4,5\}$ and two bike lane construction plans, namely A and B: plan A builds bike lanes on $\{1,2,4,5\}$ and plan B builds on $\{1,2,3,5\}$. Under the AC utility function, the cyclist's utility from both plans are the same: $4 + 2\lambda$. However, plan B may be more preferable to the cyclist if the marginal benefit from the continuity is increasing in the size of continuous bike lanes.
\end{example}

Hence, in this section, we discuss a more general class of cyclist utility functions with the consideration of continuity beyond adjacency. Given a trajectory $r$ and a bike lane construction plan $x$, let $S_x(r)$ denote the set of subsets of maximal continuous road segments with bike lanes on $r$. For instance, if $r = \{1,2,3,4,5\}$ and bike lanes are constructed on $\{1,3,4,5\}$ ($x_1=x_3=x_4=x_5=1$ and $x_2=0$), then $S_x(r)= \{\{1\},\{3,4,5\}\}$. We define a general utility function as
\begin{align}
v_x(r) = \sum_{s\in S_x(r)} f(|s|), \label{eq:genuti}
\end{align}
where $f(\cdot)$ is an increasing function. Under the adjacency-continuity utility function, $f(|s|) = |s| + \lambda(|s| - 1) = (\lambda+1)|s| -\lambda$, which is a linear function of $|s|$. And the score function used by \cite{bao2017planning} is a special case of (\ref{eq:genuti}), wherein $f(|s|) = |s|\alpha^{|s|}$ with $\alpha\geq 1$. We will refer the bike lane planning model with the general utility function as BL-GU. Although maximizing the general utility function is often challenging due to the nonlinearity, we can show that the utility function (\ref{eq:genuti}) has a desirable structure when $f(\cdot)$ is further assumed to be convex. 
\begin{theorem}\label{th:1}
	If $f(\cdot)$ is an increasing convex function, $v_x(r)$ defined in (\ref{eq:genuti}) is supermodular.
\end{theorem}

The convex assumption of $f(\cdot)$ is consistent with the notion that cyclists receive additionally more benefits by riding through more continuous bike lanes. For example, utility functions such as the ones with $f(|s|)=|s|\alpha^{|s|}$ ($\alpha>1$) are supermodular. With supermodular utility functions, the general utility maximization problem over the budget constraint has a polynomial-time solvable Lagrangian dual problem.
\color{black}

\begin{corollary}\label{cor:lagdual}
	If $f(\cdot)$ is an increasing convex function, maximizing the utility function  $v_x(r)$ defined in (\ref{eq:genuti}) over a budget constraint yields a polynomial-time solvable Lagrangian dual.
\end{corollary}

 However, different from the adjacency-continuity utility function, each iteration of the Lagrangian relaxation under the general utility function is not equivalent to a maximum flow problem. Instead, we can use a general supermodular maximization oracle such as Fujishige's minimum-norm-point algorithm \citep{fujishige2005submodular}. Then the computational performance of solving the Lagrangian dual problem heavily depends on the efficiency of the supermodular maximization oracle. In our case, the minimum-norm-point algorithm is not computationally efficient. 

Nevertheless, the bike lane planning problem using the general utility function (\ref{eq:genuti}) can be formulated as an MILP.

\begin{proposition}\label{prop:genmilp}
	Under the general utility function (\ref{eq:genuti}), the bike lane planning problem can be solved as an MILP. 
\end{proposition}
Note that the MILP formulation is attainable without assuming $f(\cdot)$ is convex. Specifically, the general utility function can be represented as 
\begin{equation}
v_x(r) = \sum_{l\in L(r)} \beta_l \prod_{i\in l} x_i  \label{eqn:generalu}
\end{equation}
with properly chosen coefficients $\beta_l$, where $L(r)$ is defined to include all the possible subsets of continuous road segments on $r$. Each element $l\in L$ is a subset of continuous road segments. Here $L(r)$ is different from $S_x(r)$ in the sense that $L(r)$ is independent of $x$. For instance, given $r=\{1,2,3\}$, then $L(r) = \{\{1\},\{2\},\{3\},\{1,2\}, \{2,3\}, \{1,2,3\} \}$ and $v_x(r) = \beta_1 x_1 + \beta_2 x_2 + \beta_3 x_3 + \beta_{1,2} x_1x_2 + \beta_{2,3}x_2x_3 + \beta_{1,2,3}x_1x_2x_3$, where coefficients $\beta$'s can be calculated as
\begin{align*}
&\beta_1 = \beta_2 = \beta_3 = f(1)),\\
&\beta_{1,2} = \beta_{2,3} = f(2) - 2f(1),\\
&\beta_{1,2,3} = f(3) - 2(f(2) - 2f(1)) - 3f(1) = f(3) - 2f(2) + f(1).
\end{align*}
More generally, $\beta_l = f(|l|) - 2f(|l|-1) + f(|l|-2)$ for a nonempty $l$ (the proper definition requires $f(-1) = 0$). When $f(\cdot)$ is an increasing convex function, all $\beta$'s are nonnegative.

\hl{The above analysis can be extended to consider heterogeneous road lengths. When the utility depends on the riding distance, the $|\cdot|$ function in (\ref{eq:genuti}) can be interpreted as the total length function for a given subset of continuous road segments. Now denote the trajectory $r$ as $\{i^1, \dots, i^n\}$, the coefficients $\beta_l$ in Equation (\ref{eqn:generalu}) can be calculated as follows: (i) When $l =\{i\}$, $\beta_l = f(|l|) = f(|\{i\}|)$; (ii) When $l = \{i^j, i^{j+1}\}$, $\beta_l = f(|l|) - f(|\{i^j\}|) - f(|\{i^{j+1}\}|) = f(|\{i^j, i^{j+1}\}|) - f(|\{i^j\}|) - f(|\{i^{j+1}\}|)$; (iii) When $l = \{i^j, i^{j+1}, \dots, i^{j+k}\}$ ($k\geq 2$), $\beta_l = f(|l|) - f(|\{i^j, i^{j+1}, \dots, i^{j+k-1}\}|) - f(|\{i^{j+1}, i^{j+2}, \dots, i^{j+k}\}|) + f(|\{i^{j+1}, \dots, i^{j+k-1}\}|)$. Hence, in the example where $r=\{1,2,3\}$, we would have
\begin{align*}
&\beta_1 = f(|\{1\}|),\ \beta_2 =f(|\{2\}|),\ \beta_3 = f(|\{3\}|),\\
&\beta_{1,2} = f(|\{1,2\}|) - f(|\{1\}|) - f(|\{2\}|),\  \beta_{2,3} = f(|\{2,3\}|) - f(|\{2\}|) - f(|\{3\}|),\\
&\beta_{1,2,3} = f(|\{1,2,3\}|) - f(|\{1,2\}|) - f(|\{2,3\}|) + f(|\{2\}|).
\end{align*}
We provide an inductive proof about the validity of the above calculation in Appendix \ref{appen:beta}.
}

Since function (\ref{eqn:generalu}) only involves product terms with binary variables, we can linearize them to get an MILP in a similar manner to \ref{obj:form2}. We call this formulation BL-GU-MILP.
\begin{align}
\tag{BL-GU-MILP}
\max_{x}\quad &  \sum_{m\in M}\sum_{l\in L(r_m)} \beta_l y_l, \label{obj:form3}\\
\mbox{s.t.} \quad & y_l \geq \sum_{i\in l} x_i - (|l| - 1), \quad \forall l\in L(r_m), m\in M, \label{const:gen1}\\
&y_l\leq x_i, \quad  \forall i\in l, l\in L(r_m), m\in M ,\label{const:gen2}\\
\quad & \sum_{i\in V} c_ix_i \leq B, \label{budget:form3}\\
& 0\leq y_l\leq 1,\quad \forall l\in L(r_m), m\in M, \label{const:gen4}\\
& x_{i}\in\{0,1\},\quad \forall i\in V.
\end{align}
Again, constraints (\ref{const:gen1}) are not necessary if $\beta$'s are nonnegative. The number of constraints (\ref{const:gen2}) may be intimidating, but we can get a simple reduction by utilizing the nested structure of $l$. For $l\in L(r)$ with $|l|>2$, two subsets, $l_-$ and $l^-$, can be obtained by removing the first and the last road segment of $l$, respectively. Since $l_-, l^- \in L(r)$, we can use two constraints $y_l \leq y_{l_-}$ and $y_l\leq y_{l^-}$ instead of constraints (\ref{const:gen2}). For example, given $l=\{1,2,3,4\}$, we can use $l_- = \{2,3,4\}$ and $l^- = \{1,2,3\}$. After the reduction, the constraint matrix of BL-GU-MILP boils down to a totally unimodular matrix along with a budget  constraint, as formalized in the following proposition.
\begin{proposition}\label{prop:milptu}
	If $f(\cdot)$ is an increasing convex function, BL-GU-MILP with the relaxed budget constraint has a totally unimodular constraint matrix. Then the corresponding Lagrangian relaxation can be solved as a linear program (LP).
\end{proposition}
Proposition \ref{prop:milptu} has several important implications. First, it presents an alternative way to prove the polynomial-time solvability result of the Lagrangian dual in Corollary \ref{cor:lagdual}. Since LP is polynomial-time solvable and the Lagrangial dual has a limited number of break points, the Lagrangian dual can be solved in polynomial time. Second, instead of resorting to a general supermodular maximization oracle, the Lagrangian dual can now be solved with a linear programming solver, which tends to deliver significantly better computational performance.

Regarding the size of BL-GU-MILP, both the number of continuous variables and the number of constraints in BL-GU-MILP are $O(\min\{|M|N_e^2, |L(V)|\})$, where $N_e$ is the size of the longest trajectory (in terms of the number of road segments). Although $|M|$ can be arbitrarily large, $L(V)$, the set of all possible continuous road segments on the entire road network, is limited. And in practice, many trajectories are similar so the actual number of variables and constraints is limited. 



\subsubsection{A Lagrangian Relaxation Based Algorithm.}\label{subsec:lag} Solving BL-GU-MILP directly via commercial MILP solvers can be challenging due to the large number of variables and constraints in practical applications. Given that the Lagrangian relaxation subproblem of BL-GU-MILP by relaxing the budget constraint can be solved as an LP, we propose a simple and efficient Lagrangian relaxation based algorithm (Algorithm \ref{alg:lag}) to solve the large-scale bike lane planning problem under the general supermodular utility functions. \hl{Following the notations in Subsection \ref{sec:ac}, $u$ is the dual variable of the budget constraint and $\Phi(u)$ is the dual objective function. For the ease of exposition, we use $S(x)$ to denote the objective function of BL-GU-MILP, $c(x)$ for the construction cost ($=\sum_{i\in V}c_i x_i$),  $x(u)$ for the maximizer of the Lagrangian relaxation at $u$, and $g(u) = -(c(x) - B)$ for the subgradient. Furthermore, let $e=(1, \dots, 1)\in \mathbb{R}^{|V|}$ denote the vector of all ones, which corresponds to the decision of constructing bike lanes on all road segments.}

\begin{algorithm}[!h]
\caption{{\em GU-Lag: Lagrangian Relaxation Heuristic for BL-GU-MILP}} \label{alg:lag}
	\quad \textbf{Input}: tolerance level $\epsilon >0$.
\begin{algorithmic}[1]
		\STATE Initialize $u'=0$ and $u''=\max_i \frac{S(e)}{c_i}$. Correspondingly, $x(u')=e$, $x(u'') = 0$, and so $g(u') = -(c(e) - B)$ and $g(u'') = B$.
		\STATE  Calculate $u^* = \frac{S(x(u'')) - S(x(u'))}{c(x(u'')) - c(x(u'))}$, and solve for $x(u^*)$ with LP. \label{line:2}
		\WHILE{$|\Phi(u^*) - \left(\Phi(u'') + (u^* - u'')g(u'')\right)|/\Phi(u^*) > \epsilon$} \label{line:3}
			\STATE Set $u''=u^*$ if $c(x(u^*))>B$ and $u'= u^*$ otherwise; \label{line:5}
			\STATE  Update $u^* = \frac{S(x(u'')) - S(x(u'))}{c(x(u'')) - c(x(u'))}$, and solve for $x(u^*)$ with LP.
		\ENDWHILE \label{line:7}
		\IF{$c(x(u^*))= B$} \label{line:8}
		\STATE return $x(u^*)$;
		\ELSE
		\STATE Set up a new BL-GU-MILP by removing variables $x_i$ for $i\in\{j\in V: x_j(u^*)=0\}$ and $y_l$ for $l\in \{l'\in \{L(r_m)\}_{m \in M}: \exists i\in l', x_i(u^*)=0 \}$; solve this BL-GU-MILP and return the solution. \label{line:11}
		\ENDIF
\end{algorithmic}
\end{algorithm}

\hl{Algorithm \ref{alg:lag} consists of two main parts. In the first part (steps \ref{line:2}-\ref{line:7}), we solve the Lagrangian dual of BL-GU-MILP (after relaxing the budget constraint) with an outer approximation algorithm. The outer approximation algorithm maintains a lower limit $u'$ and an upper limit $u''$ for the optimal dual solution such that $g(u') <0$ and $g(u'')\geq 0$. Then the function $\Psi^{u', u''}(u) = \max\{\Phi(u') + (u - u')g(u'), \Phi(u'') + (u - u'')g(u'')\}$ is an outer approximation of $\Phi(u)$ (note that $\Phi(\cdot)$ is a convex piecewise linear function). Observe that $\Psi^{u', u''}(u)$ has a minimizer of $u^* = [(\Phi(u'') -  u''g(u'')) - (\Phi(u') -  u''g(u'))  ]/(g(u') - g(u'')) =  [S(x(u'')) - S(x(u'))]/(c(x(u'')) - c(x(u'))) $, and we use it to update the lower/upper limit.  Because $\Phi(u)$ has at most $|V|$ break points, this outer approximation algorithm terminates with at most $|V|+1$ iterations \citep{gallo1989supermodular}. At termination, we obtain $u^*$ and $x(u^*)$, which are the optimal dual solution and its corresponding primal solution, respectively. If the solution  $x(u^*)$ satisfies the budget constraint at equality, then it is also optimal to BL-GU-MILP; otherwise we select road segments from the subset $V(u^*) = \{i\in V: x_i(u^*) = 1\}$ by solving a new MILP, as presented in step \ref{line:11} of the algorithm. Because $V(u^*)$ is smaller than $V$ (especially when $B$ is not so large), the new MILP involves much less variables and can be solved more efficiently. We test the efficiency of this algorithm in the numerical study.}


\subsection{Bike Lane Planning with Route Choices}\label{subsec:behav}
Up till now our model makes no assumptions about cyclists' responsive behaviors to the bike lane construction plan. The aforementioned models maximize the cyclists' utility assuming their route choices are fixed (their trajectories will not be impacted by the constructed bike lanes). This assumption may not be valid if cyclists update their route choices based on the constructed bike lanes. Since the utility from riding through a route is influenced by the constructed bike lanes, cyclists may choose to take a different route than the observed trajectory if more bike lanes are constructed along that route. 

To account for cyclists' route choices, we assume for a cyclist $m$ riding from $i\in V$ to $j\in V$, she can choose from a set of candidate routes/trajectories $C_{m}=\{r_{m}^1,\dots, r_{m}^{t_{m}}\}$, where each route starts with $i$ and ends with $j$. In addition to the bike lane utility, cyclists' evaluation of a route also depends on the travel time, slope, noises, and other physical characteristics. Therefore, we add to the utility function an exogenous disutility term $\bar{v}(r)$ that mainly captures the travel time cost of route $r$. In practice, $\bar{v}(r)$ can be estimated beforehand and assumed to be known.

Given a bike lane construction plan $x$, cyclist $m$ chooses the route $r\in C_m$ with probability $p_{mr}$ according to some choice model. \hl{Then the objective of the bike lane planning problem is
\begin{equation}\label{eqn:obj_choice}
\max_x \quad \sum_{m\in M} \sum_{r\in C_m} D_m p_{mr} (v_x(r) - \bar{v}(r)),
\end{equation}
where $D_m$ is the aggregated demand of cyclist $m$ (after aggregation based on origin-destinations).}

The bike route choice is often modeled using the Multinomial Logit model (MNL), as shown in \cite{hood2011gps} and \cite{khatri2016modeling}. Under the MNL model, the probability $p_{mr}$ is given by
\begin{equation}\label{eqn:choice}
p_{mr} = \frac{\exp(v_x(r) - \bar{v}(r))}{  \sum_{r'\in C_m} \exp(v_x(r') - \bar{v}(r'))}.
\end{equation}
Because many alternative routes have overlapping road segments, the assumption of irrelevant alternatives may not be satisfied. To relieve this concern, a correction term called Path Size factor (PS) can be added, e.g., see \cite{broach2012cyclists}. This correction term can be calculated beforehand and is independent of the bike lane decision, thus the structure of $p_{mr}$ remains the same.

The consequent bike planning problem is similar to an assortment optimization problem as decisions in both problems shape the choice probabilities. However, while the assortment decision influences the choice probabilities by altering the choice set, the bike lane construction decision transforms the choice probabilities by changing the utility values. That being said, the choice set in our problem is invariant to the decision variables, which is a critical difference between our problem and the assortment optimization problem. Furthermore, unlike the assortment optimization where the marginal profit of each product is exogenously given, the ``profit" from bike lanes $v_x(r)$ depends on the decision variables. Hence problem (\ref{eqn:obj_choice}) also shares a similar structure to the joint assortment-pricing optimization problem. However, the decisions here are binary and the analysis in the pricing literature can not carry on.

\hl{
To tackle the above challenges, we propose to solve the bike lane planning problem with cyclists' route choices as a bilevel program, where cyclists are viewed as followers who make route choices after observing the bike lane decisions. Equation (\ref{eqn:choice}) can be viewed as the solution to a lower-level (follower) equilibrium problem. Specifically, this equation can be derived by solving the following convex optimization problem:
\begin{align}
\tag{LL}
\min_{p_{mr}:m\in M, r\in C_m}\quad &\sum_{m\in M} \sum_{r\in C_m}\big[ p_{mr}\ln p_{mr} + p_{mr}\left(\bar{v}(r) - v_x(r) \right)  \big]\\
\mbox{s.t.} \quad & \sum_{r \in C_m} p_{mr} = 1,\ \forall m\in M \label{const:lldemand}\\
& p_{mr} \geq 0,\ \forall m\in M, r \in C_m.
\end{align}
To see this, first note that the optimal solution to (LL) must be positive, then let $\gamma$ denote the dual variable of constraint (\ref{const:lldemand}). The optimality condition is
\[  \ln p_{mr} + 1 + \bar{v}(r) - v_x(r) = \gamma_m,\ \forall r \in C_m, \]
which gives $p_{mr} = \exp(v_x(r) - \bar{v}(r))/\exp(1 - \gamma_m)$. Combining with Equation (\ref{const:lldemand}), we obtain that
\[ p_{mr} =   \frac{\exp(v_x(r) - \bar{v}(r))}{  \sum_{r'\in C_m} \exp(v_x(r') - \bar{v}(r'))}.     \]
Therefore, the original bike lane planning model can be reformulated as a bilevel program with the lower-level problem (LL). Because (LL) is convex, we can borrow the ideas from convex bilevel programming to solve the resulting bike lane planning model. 
}

\hl{
Motivated by \cite{dan2019competitive}, we devise a linear approximation approach that approximates the convex parts in the objective function of (LL) by piecewise linear functions. Specifically, we sample from $[0,1]$ $K$ points $\{p^1, p^2, \dots, p^K\}$ such that $p^i<p^j$ for all $i<j$. Then the piecewise linear approximation of $p_{mr}\ln p_{mr}$ takes the form of
 \[  p(\ln p^k + 1) - p^k,\quad \forall k=1,\dots, K.        \]
Therefore, we obtain the linear approximation of (LL) as 
\begin{align}
\tag{LL-Lin}
\min_{p_{mr}: m\in M, r\in C_m}\quad & \sum_{m\in M}\sum_{r\in C_m}\big[ \omega_{mr} + p_{mr}\left(\bar{v}(r) - v_x(r) \right)  \big]\\
\mbox{s.t.} \quad & \sum_{r \in C_m} p_{mr} = 1,\ \forall m \in M \label{constr:lin1}\\
&\omega_{mr} \geq p_{mr}(\ln p^k + 1) - p^k,\quad \forall  m\in M, r\in C_m, k=1,\dots, K, \label{constr:lin2}\\
& p_{mr} \geq 0,\ \forall m\in M, r \in C_m. \label{constr:lin3}
\end{align}
}
\hl{Next, we can replace (LL) in the bike lane planning problem by a set of optimality condition constraints associated with (LL-Lin), i.e., primal/dual feasibility and complementary slackness constraints. In order to derive an MILP, the final step is to linearize product terms $p_{mr}v_x(r)$ that appear in both (LL-Lin) and the objective function (\ref{eqn:obj_choice}). Note that $v_x(r)$ is a sum of binary variables, and we can write $p_{mr}v_x(r)$ as
\[ p_{mr}v_x(r) = p_{mr}\sum_{l\in L(r)} \beta_ly_l =  \sum_{l\in L(r)} \beta_l \zeta_{mrl},  \]
where auxiliary variables $\zeta_{mrl}$ are introduced along with the following two linear constraints
\begin{align*}
&\zeta_{mrl} \leq p_{mr},\quad \forall l\in L(r),\\
&\zeta_{mrl} \leq y_l,\quad \forall l\in L(r).
\end{align*}
Now the bike lane planning problem with cyclists' route choices has been transformed into an MILP. We provide the detailed formulation and more technical details in Appendix \ref{appen:lin}
}

\hl{
The performance of our linear approximation approach depends on the linear approximation accuracy. We prove that this approach is asymptotically exact when $K$ increases, as indicated in the following result.
\begin{proposition}\label{prop:approx}
	The approximation error of the objective function (\ref{eqn:obj_choice}) with (LL-Lin) is $O(\frac{1}{K})$.
\end{proposition}
}

\section{A Real-World Case Study}\label{sec:case}
We apply the proposed models and algorithms to a real-world trajectory data set. \hl{First, we describe the data set and present its summary statistics. Then we discuss the computational performance and solution quality of the proposed algorithms. Lastly, we compare the bike lane construction plans generated by different models with varying parameters, with and without consideration of cyclists' route choice models.}

\subsection{Data Description} \label{sec:data}
We obtain a GPS bike trajectory data set via our collaboration with the urban planning institution of Zhuhai city and a major bike-sharing company operating in Zhuhai. Zhuhai is a medium-size city of China with a population of 1.67 million, where both station-based and dock-less bike-sharing systems have been deployed. 
\hl{We collected bike trajectories in the central urban area of Zhuhai from our partner bike-sharing company from March 2017 until February 2018. Between Aug 2017 and Feb 2018, the ridership shrank significantly as the company shifted their operations focus. Therefore, we decided to only keep the data from March-July 2017.  Note that during the collection period, there were two other dock-less bike-sharing platforms operating in Zhuhai, and the demand was split between these platforms. Since the competing platforms share a similar business models (and their bikes have similar designs), we assume our trajectory data can approximately represent the travel pattern and route preferences of bike riders.} 

\hl{Nevertheless, the data set is potentially biased due to bike availability and infrastructure issues. The observed demand is censored as potential ride demand may be lost due to the lack of available bikes or safe bike infrastructure in the neighborhood. To alleviate the the availability bias, we apply a decensoring approach to reconstruct the true demand, which will be detailed in the sequel. To uncover the lost demand due to missing infrastructure, we can potentially combine the observed demand data with user click data (on mobile apps) and survey data to understand the riders' behaviors and preferences (on safe infrastructure). We will leave this as a future research direction.}

The obtained trajectories are mapped to the road network extracted from OpenStreetMap \citep{Osm2018}\footnote{ To map the GPS coordinates to the road network, we generate a near table in ArcGIS that finds the nearest road segment to a coordinate point. Then each coordinate is associated with a road segment ID.}. Each trajectory contains a timestamp that indicates the start time of a trip and a series of GPS coordinates that were recorded every 5 seconds. After removing trajectories that are shorter than one minute, there are \hl{109,640} bike trajectories in total. 


\hl{
\textbf{Data Decensoring} Our original trajectory data is subject to the censoring bias from stock-out events, e.g., potential riders can not fulfill their demand if the neighborhood area has no bikes available. To address this issue, we follow the idea from \cite{o2015data} to decensor the trip demand and the corresponding trajectory data. Specifically, we first identify the stock-out events from the observed stock evolution in each neighborhood, 
and then calculate the average demand conditional on that there are no stock-outs for each neighborhood. Afterwards we reweight the trajectory data to be aligned with the conditional average demand. We provide additional details about this decensoring procedure in Appendix \ref{appen:decen}.
}


\textbf{Summary Statistics and Spatial Distribution} Figure 1 presents the distribution of bike trajectory duration from our data set. The average trajectory duration is 903.6 seconds (15.1 minutes),  
and the majority of trajectories have duration shorter than 20 minutes. This is because most trajectories are limited to the urban and residential areas of Zhuhai. 
We identify 3,735 road segments from the trajectories in total. The average length of the road segments is 195.3 m, and more than 60\% of road segments have lengths under 200 meters. Most roads are constructed in the urban area of Zhuhai with high density of population, and thus intersections are close to each other in this area.

\begin{figure}[htbp]
	\begin{center}
		\includegraphics[width=0.4\textwidth]{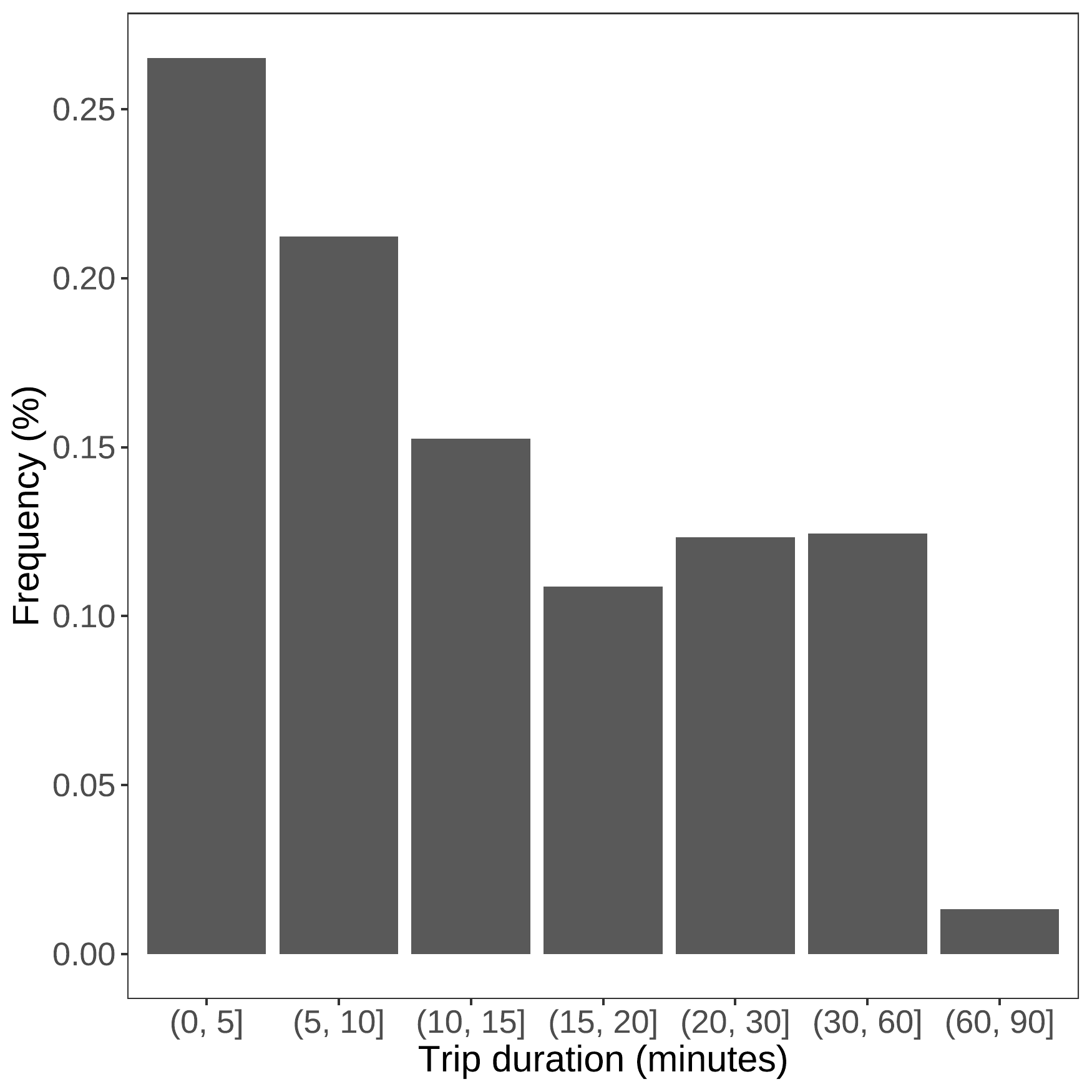}
		\caption{Bike trajectory duration distribution. }\label{fig:duration}
	\end{center}
\end{figure}

\begin{figure}[htbp]
	\begin{center}
		\includegraphics[width=0.7\textwidth,trim={0cm 0cm 0cm 0cm}, clip]{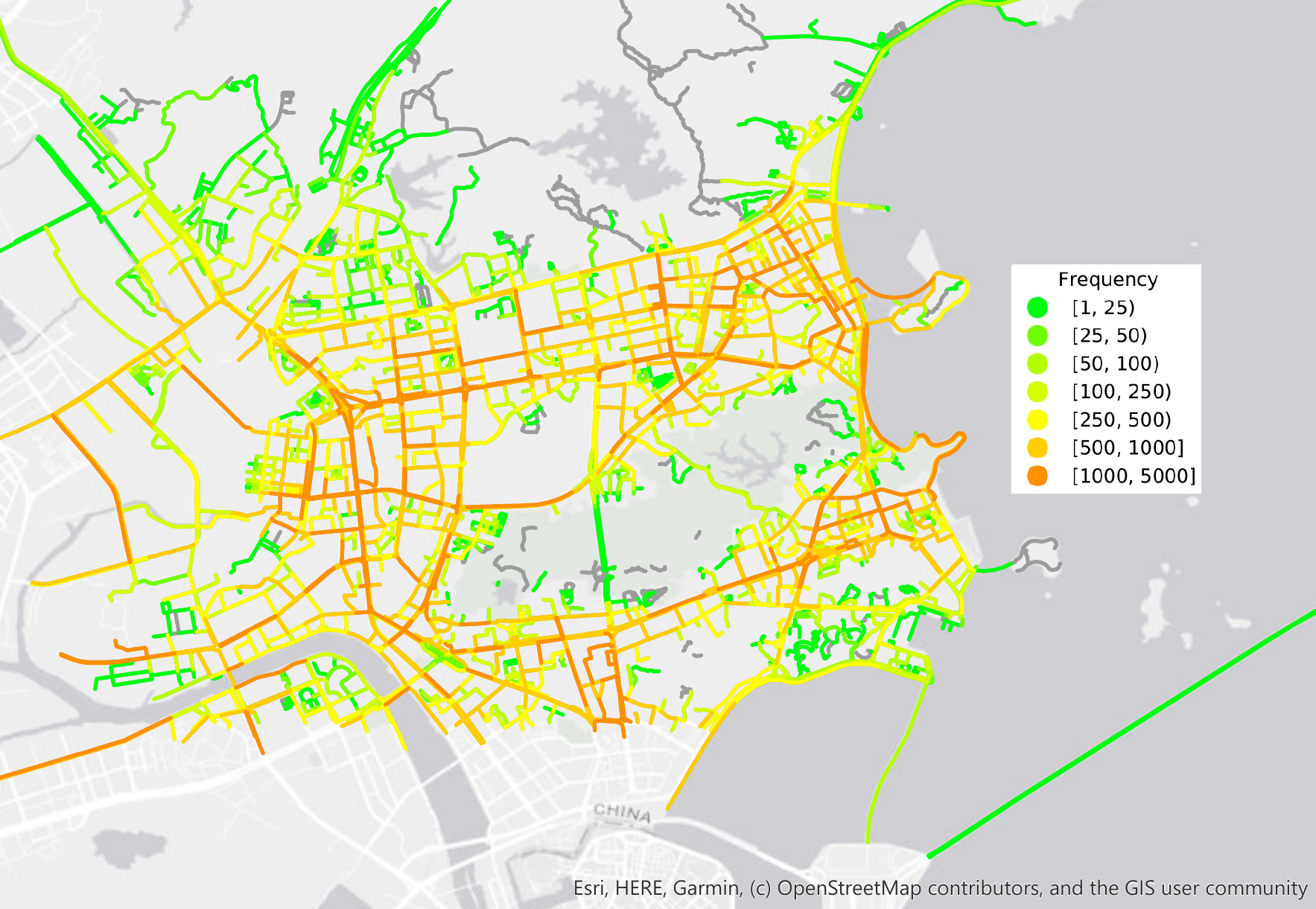}
		\caption{Road segment usage spatial distribution.}\label{fig:usagespatial}
	\end{center}
\end{figure}

Figure \ref{fig:usagespatial} presents the spatial distribution of the road segment usage frequency (the number of trajectories passing through the road segment). We observe that the locations of the road segments with high usage levels are located in a few areas in Zhuhai, which include the financial district, shopping district, and residential districts of the city. We also find that many popular road segments are spread out over the city, which implies that merely maximizing the coverage of bike trips would result in a highly discontinuous bike lane system. More details about the data set can be found in Appendix \ref{appen:temp}.

\subsection{Computational Result} \label{subsec:comp}
We evaluate the computational performance of the proposed models and algorithms on the practical road network and trajectory data set.  For BL-GU, the utility function takes the form of $v_x(r)=\sum_{s\in S_x(r)} |s|\alpha^{|s|}$. We vary the choices of $\lambda$, $\alpha$, $B$, and the number of sampled trajectories (cyclists) $m$.
The experiments were conducted with Gurobi (Gurobi Optimization 2018) and ran on a Windows 10 64-bit machine with a Intel Xeon 4114 2.20 GHz processor and 32.0
GB RAM.  

BL-AC can be solved by the MILP solver efficiently (BL-AC-MILP), and the solution time is within a minute for the studied road network. 
Different from BL-AC, the general model BL-GU often involves a much greater number of variables and solving the corresponding MILP formulation can be time consuming. We compare here the efficiency of the proposed Lagrangian relaxation based algorithm (denoted as GU-Lag) versus a benchmark greedy heuristic adapted from \cite{bao2017planning}, which selects road segments to increase the utility function in a greedy way. The detailed comparison results are presented in Table \ref{tab:runtime_gu}, in which we test the two algorithms on different sets of trajectories (e.g., $m=2000$ indicates a random sample of $2000$ trajectories). \hl{The reported optimality gap is $(Z^{UB} - Z)/Z^{UB}$, where $Z^{UB}$ is the upper bound derived from the Lagrangian dual $\Phi(u^*)$ and $Z$ is the objective value of the feasible solution found by the algorithm.}
The reported performance is averaged across five different runs. We observe that GU-Lag delivers superior performance in terms of both the running time and solution quality. In particular, the greedy algorithm does not scale well to the cases with large values of $B$ or $\alpha$. For example, when $\alpha=1.1$, the solution derived from the greedy algorithm has a relatively large optimality gap. By contrast, GU-Lag admits a reliable computational performance across all different combinations of parameters. 
\begin{table}[htbp]
	\small
	\centering
	\caption{Computational performance of GU-Lag and the greedy algorithm (time in seconds)}
	\begin{tabular}{cccccccccccccc}
		\toprule
		&       & \multicolumn{4}{c}{$m=2000$}   & \multicolumn{4}{c}{$m=4000$}   & \multicolumn{4}{c}{all trajectories} \\
		\cmidrule[1.2pt](lr{1em}){3-6}\cmidrule[1.2pt](lr{1em}){7-10}\cmidrule[1.2pt](lr{1em}){11-14}
		\multicolumn{1}{c}{\multirow{2}[0]{*}{$B$}} & \multicolumn{1}{c}{\multirow{2}[0]{*}{$\alpha$}} & \multicolumn{2}{c}{GU-Lag} & \multicolumn{2}{c}{Greedy} & \multicolumn{2}{c}{GU-Lag} & \multicolumn{2}{c}{Greedy} & \multicolumn{2}{c}{GU-Lag} & \multicolumn{2}{c}{Greedy} \\
		\multicolumn{1}{c}{(km)} & \multicolumn{1}{c}{} & Time  & Gap   & Time  & Gap   & Time  & Gap   & Time  & Gap   & Time  & Gap   & Time  & Gap \\
		\midrule
		\multirow{3}[0]{*}{30}  & 1.02        & 28.3     & 0.03\%   & 156      & 17.26\%  & 63.2     & 0.03\%   & 382      & 17.96\%  & 111      & 0.02\%   & 767      & 14.47\% \\
		& 1.05         & 44.1     & 0.64\%   & 130      & 17.98\%  & 216      & 0.70\%   & 317      & 21.00\%  & 1,046    & 0.67\%   & 664      & 14.06\% \\
		& 1.1       & 35.4     & 1.64\%   & 106      & 58.57\%  & 277      & 1.59\%   & 257      & 74.51\%  & 243      & 2.63\%   & 440      & 70.93\% \\
		\midrule
		\multirow{3}[0]{*}{50}  & 1.02      & 28.5     & 0.02\%   & 437      & 13.69\%  & 65.5     & 0.03\%   & 1,007    & 13.08\%  & 118      & 0.02\%   & 1,893    & 12.72\% \\
		& 1.05     & 48.7     & 0.24\%   & 360      & 19.16\%  & 249      & 0.62\%   & 782      & 21.19\%  & 361      & 0.69\%   & 1,545    & 11.01\% \\
		 & 1.1      & 58.7     & 0.86\%   & 247      & 45.30\%  & 146      & 0.93\%   & 602      & 59.64\%  & 2,449    & 2.01\%   & 945      & 50.28\% \\
		\midrule
		\multirow{3}[0]{*}{100}  & 1.02      & 35.0     & 0.03\%   & 1,810    & 8.15\%   & 78.6     & 0.02\%   & 4,560    & 5.13\%   & 164      & 0.03\%   & 8,626    & 2.97\% \\
		& 1.05      & 106      & 0.05\%   & 1,462    & 15.14\%  & 207      & 0.04\%   & 3,318    & 16.90\%  & 412      & 0.02\%   & 7,167    & 14.40\% \\
		& 1.1      & 112      & 0.20\%   & 973      & 26.49\%  & 462      & 0.13\%   & 2,046    & 43.85\%  & 202      & 0.03\%   & 3,852    & 38.15\% \\
		\bottomrule
	\end{tabular}%
	\label{tab:runtime_gu}%
\end{table}%

\hl{Regarding the bike lane planning model with route choices, we implement the linear approximation scheme and the reformulated MILP as the solution approach. We set the approximation sample size $K=20$, and the termination criteria to be: 1) the optimality gap (MIP gap) is below 0.1\%; or 2) the solution time exceeds 6 hours. The computational performance is summarized in Table \ref{tab:runtime_guchoice} (we only present the result for BL-GU as it is more difficult to solve than BL-AC in general). We observe that for practical size problems, a high-quality solution (with a small optimality gap) to the linear approximation model can be found within a few hours. The optimality gap is less than 0.1\% for most instances and is 0.2\% in the worst case. Hence the proposed MILP reformulation model is computationally tractable.}

\begin{table}[htbp]
	\small
	\centering
	\caption{Computational performance of the MILP reformulation of BL-GU with route choices (time in seconds)}
	\begin{tabular}{cccccccc}
		\toprule
		&       & \multicolumn{2}{c}{$m=2000$}   & \multicolumn{2}{c}{$m=4000$}   & \multicolumn{2}{c}{all od-pairs} \\
		\cmidrule[1.2pt](lr{1em}){3-4}\cmidrule[1.2pt](lr{1em}){5-6}\cmidrule[1.2pt](lr{1em}){7-8}
		\multicolumn{1}{c}{$B$ (km)} & \multicolumn{1}{c}{$\alpha$}  & Time  & Gap   & Time  & Gap   & Time  & Gap  \\
		\midrule
		\multirow{3}[0]{*}{30} & 1.02     & 2,060    & 0.09\%   & 5,437    & 0.09\%   & 8,299    & 0.09\% \\
		& 1.05     & 2,658    & 0.09\%   & 4,664    & 0.09\%   & 6,657    & 0.09\% \\
		& 1.1      & 6,834    & 0.07\%   & 16,761   & 0.10\%   & 21,600   & 0.12\% \\
		\midrule
		\multirow{3}[0]{*}{50} & 1.02     & 1,597    & 0.09\%   & 5,233    & 0.09\%   & 21,600   & 0.11\% \\
		& 1.05     & 2,708    & 0.09\%   & 4,182    & 0.08\%   & 14,651   & 0.07\% \\
		& 1.1      & 5,428    & 0.09\%   & 16,806   & 0.10\%   & 21,600   & 0.20\% \\
		\midrule
		\multirow{3}[0]{*}{100} & 1.02     & 586      & 0.07\%   & 3,970    & 0.08\%   & 8,328    & 0.07\% \\
		& 1.05     & 1,683    & 0.08\%   & 2,245    & 0.08\%   & 6,622    & 0.06\% \\
		& 1.1      & 3,055    & 0.08\%   & 12,970   & 0.08\%   & 19,110   & 0.09\% \\
		\bottomrule
	\end{tabular}%
	\label{tab:runtime_guchoice}%
\end{table}%

\subsection{Bike Lane Planning Result and Discussion}
We compare the bike lane planning solutions generated from BL-AC and BL-GU in terms of quantitative topological measures as well as visualization results. The setup is the same as in Subsection \ref{subsec:comp} and we use the decensored trajectories (weighted) as the model input. We first focus on the case with fixed trajectories, and then discuss the effects of incorporating the route choice model. 

\subsubsection{Topological Comparisons.} We consider five relevant topological features: the number of selected bike lanes (road segments), the number of continuous bike lane pairs, the mean number of connections per bike lane, the mean size of continuous bike lanes (along trajectories), and the max size of continuous bike lanes (along trajectories). All the features except the number of selected bike lanes measure the continuity of bike lanes on the road network. Table \ref{tab:topo} presents the comparison results based on these features for BL-AC and BL-GU with different parameter values. Note that BL-GU with $\alpha=1$ is equivalent to GL-AC with $\lambda=0$.

\begin{table}[htbp]
	\centering
	\caption{\hl{Topological comparison of BL-AC and BL-GU with $B=50$ km.}}
	\begin{tabular}{cccccc}
		\toprule
		& \multicolumn{1}{p{2.5cm}}{\centering \# of selected bike lanes} & \multicolumn{1}{p{2.8cm}}{\centering \# of continuous bike lane pairs} & \multicolumn{1}{p{2.8cm}}{\centering mean \# of connections per bike lane} & \multicolumn{1}{p{2.8cm}}{\centering mean size of continuous bike lanes} & \multicolumn{1}{p{2.8cm}}{\centering max size of continuous bike lanes} \\
		\midrule
		\multicolumn{6}{c}{BL-AC} \\
		$\lambda =0$ &  549      & 1,716    & 3.7      & 3.3      & 16 \\
		$\lambda = 2$ &  472      & 2,257    & 5.8      & 4.7      & 21 \\
		$\lambda = 10$ & 458      & 2,314    & 6.1      & 5.1      & 26 \\
		\multicolumn{6}{c}{BL-GU} \\
		$\alpha=1.02$ &  515      & 1,896    & 4.4      & 4.4      & 30 \\
		$\alpha=1.05$ & 419      & 1,916    & 5.6      & 5.9      & 53 \\
		$\alpha=1.1$ & 327      & 1,728    & 6.6      & 6.5      & 65 \\
		\bottomrule
	\end{tabular}%
	\label{tab:topo}
\end{table}%

We have several observations. First, in BL-AC (and BL-GU), increasing the value of $\lambda$ (and $\alpha$) leads to fewer selected bike lanes. Because when the continuity is assigned a small weight, the planning model tends to build bike lanes on more road segments to cover more bike trajectories. And when the continuity is more preferred, it is beneficial to build bike lanes in a few areas to make sure the bike lanes are connected to each other. Second, as $\lambda$ grows in BL-AC, the number of continuous bike lane pairs increases, which is consistent with the objective function of BL-AC that maximizes the adjacency continuity. In the meanwhile, the mean size of continuous bike lanes, the maximum size of continuous bike lanes, and the mean number of connections per bike lane also increase. By contrast, we observe that the number of continuous bike lane pairs first increases and then decreases while other continuity measures increase with $\alpha$ in BL-GU. This is because given a larger value of $\lambda$, BL-GU tends to value the size of continuous bike lanes more than the adjacency continuity. Consequently, longer trajectories will play a greater role in the planning decision and the maximum size of continuous bike lanes grows as well.    
Furthermore, the mean size of continuous bike lanes from BL-AC is often smaller than that from BL-GU. Even when $\lambda$ is very large, BL-AC can not get the same continuity level as BL-GU (in terms of the mean size of continuous bike lanes). This highlights the limitation of BL-AC, of which the objective function only considers the pair-wise continuity. BL-GU, however, measures the size of continuous bike lanes explicitly. 
 
\begin{figure}
	\subfloat[$\lambda = 0$\label{sfig:ac0}]{
		\includegraphics[width=0.5\textwidth,trim={3cm 2cm 4cm 2cm}, clip, keepaspectratio]{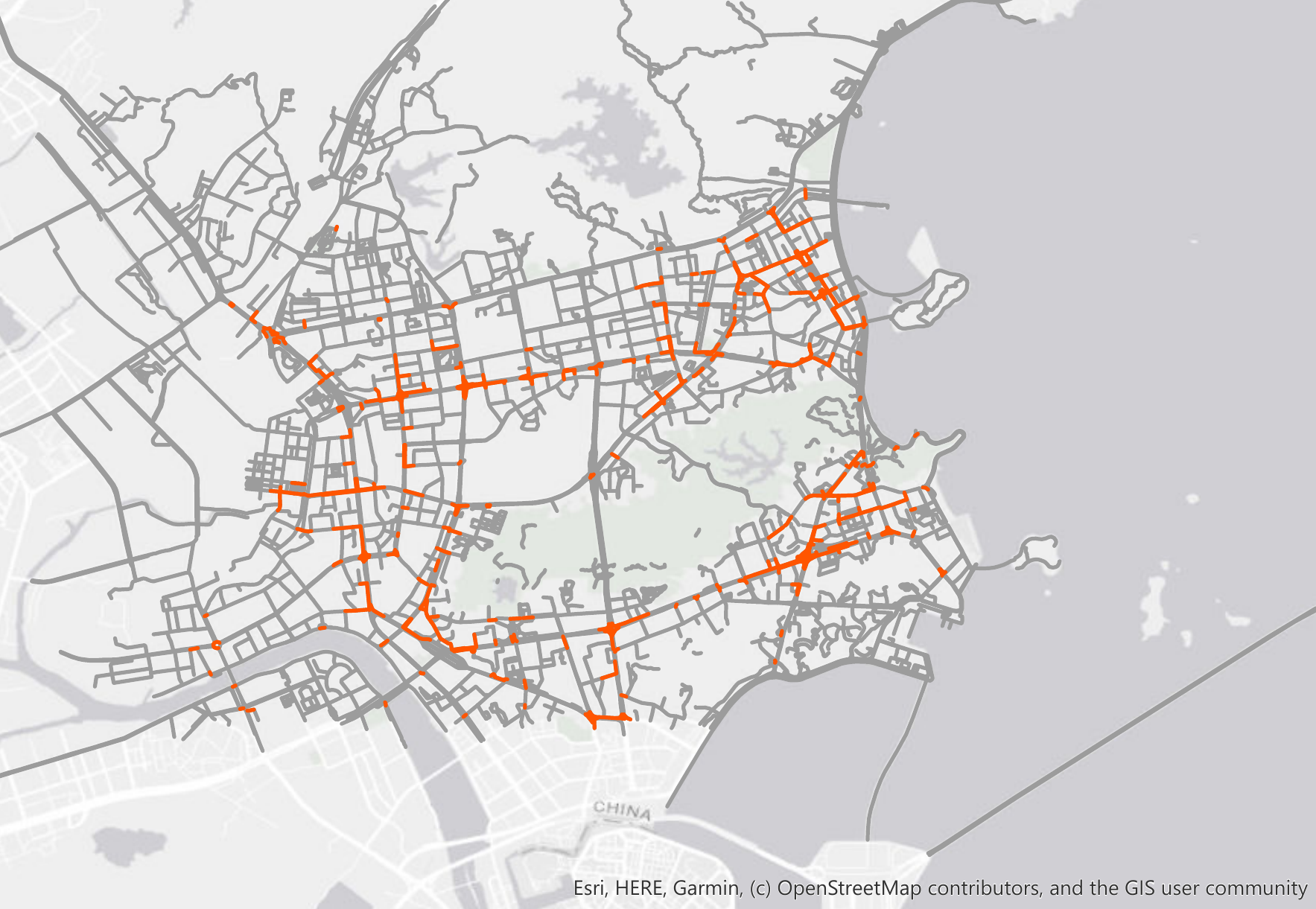}
	}\hfill
	\subfloat[$\lambda=2$ \label{sfig:ac2}]{
		\includegraphics[width=0.5\textwidth,trim={3cm 2cm 4cm 2cm}, clip, keepaspectratio]{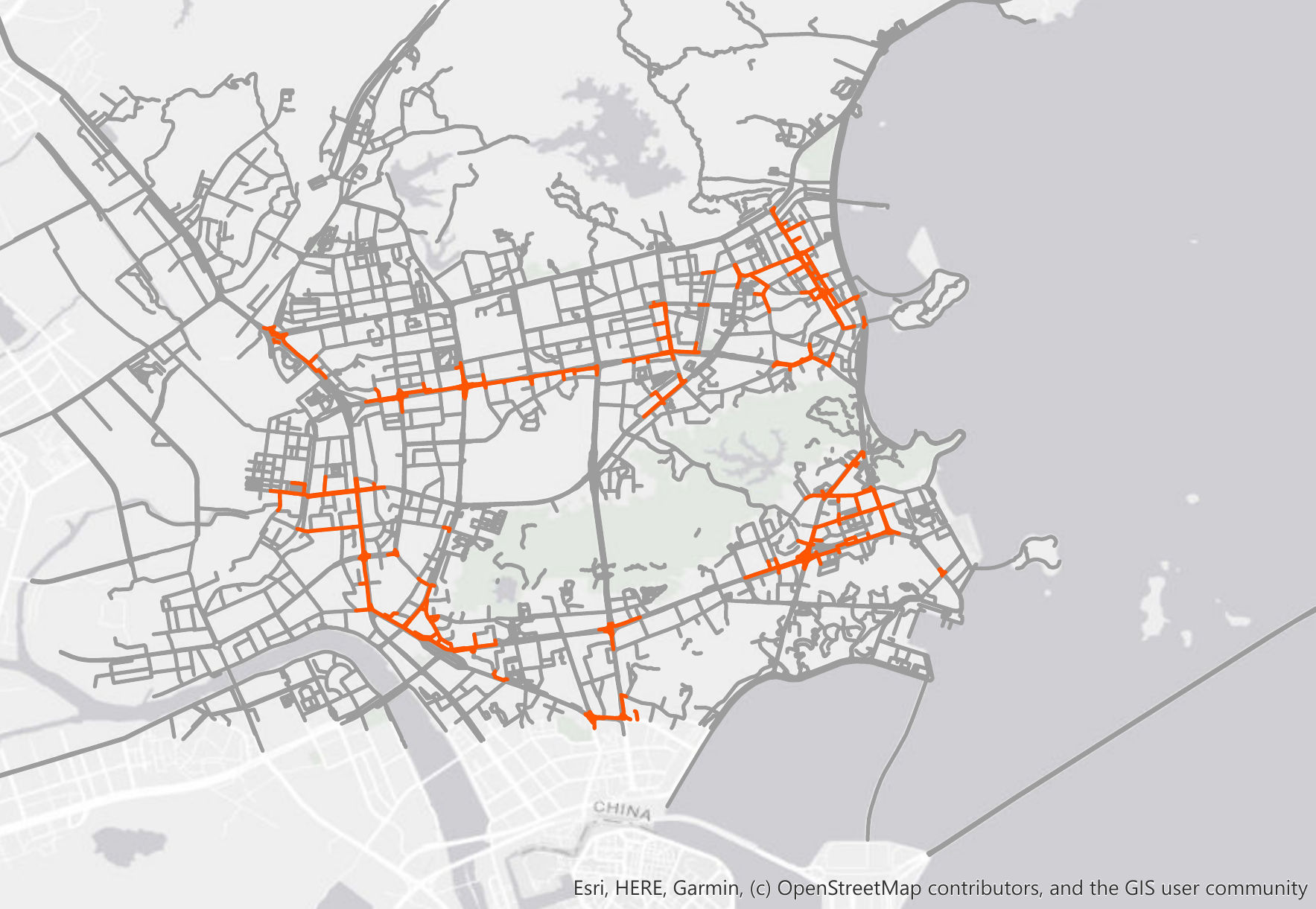}
	}\hfill
	\caption{\hl{Selected bike lanes (in red color) from BL-AC with $B=50$ km.}}\label{fig:result_ac}
\end{figure}

\begin{figure}
	\subfloat[$\alpha=1.02$\label{sfig:gu105}]{
		\includegraphics[width=0.5\textwidth,trim={3cm 2cm 4cm 2cm}, clip, keepaspectratio]{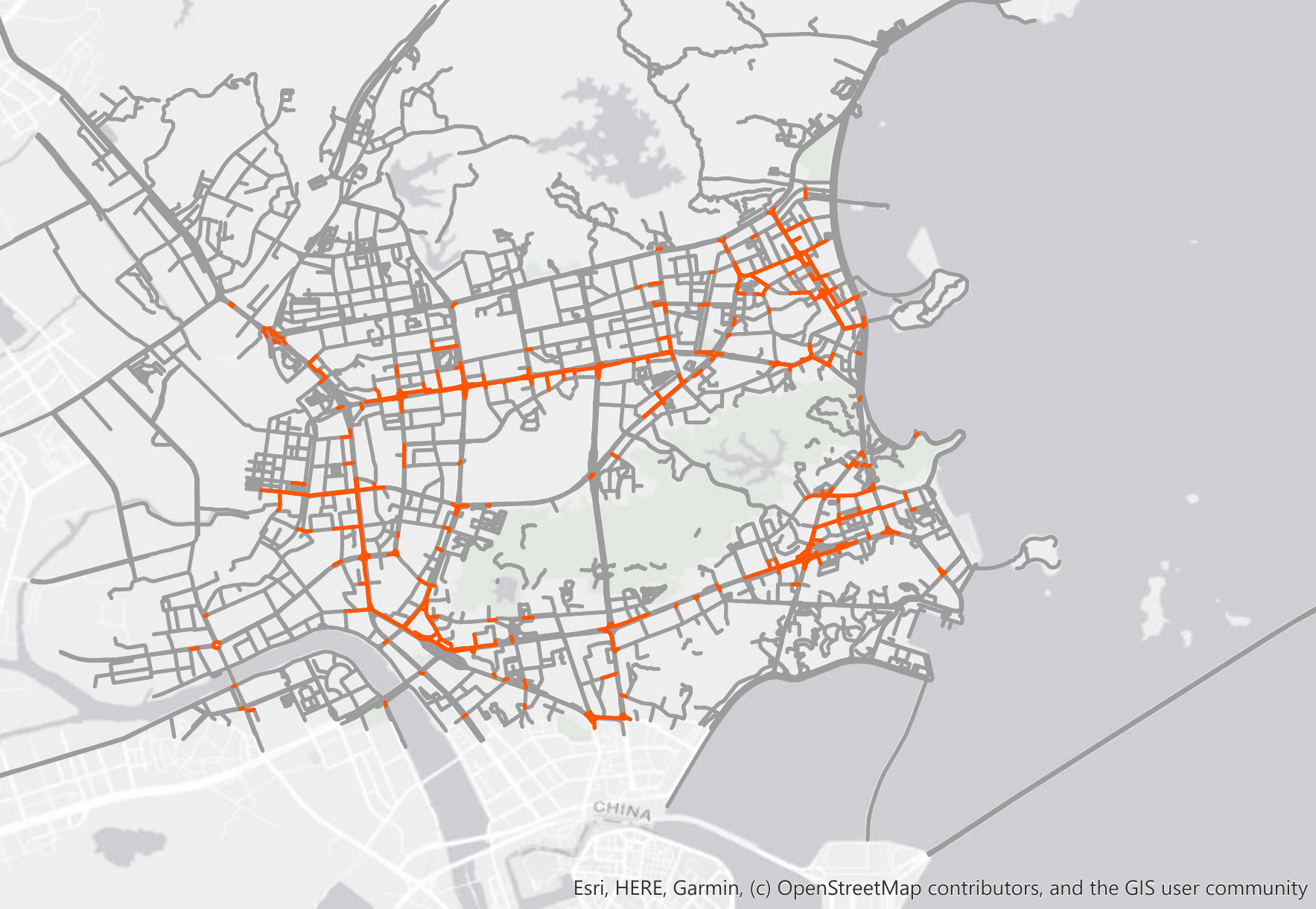}
	}\hfill
	\subfloat[$\alpha=1.05$\label{sfig:gu110}]{
		\includegraphics[width=0.5\textwidth,trim={3cm 2cm 4cm 2cm}, clip, keepaspectratio]{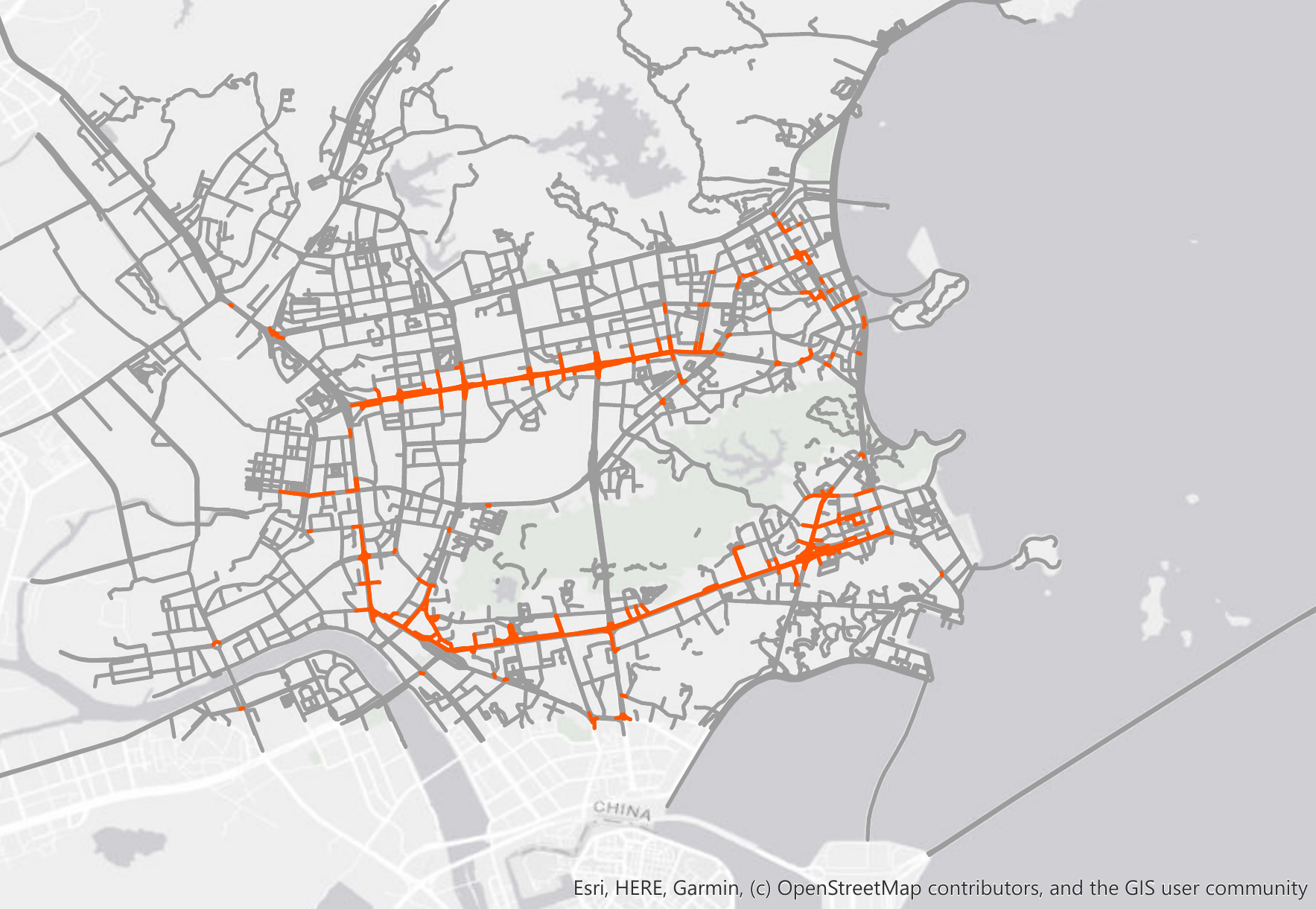}
	}\hfill
	\caption{\hl{Selected bike lanes (in red color) from BL-GU with $B=50$ km.}}\label{fig:result_gu}
\end{figure}

We visualize the bike lane planning results of BL-AC and BL-GU on the road network in Figure \ref{fig:result_ac} and Figure \ref{fig:result_gu}, respectively.  It can be shown that when $\lambda=0$ ($\alpha=1$), the selected bike lanes are spread out over the city. Consistent with our topological findings, a large value of $\lambda$ in BL-AC induces fewer and more continuous bike lanes. Increasing the value of $\alpha$ in BL-GU has a similar effect that yields a more continuous bike lane network,  as shown in Figure \ref{sfig:gu105}. 
Different from BL-AC, as indicated by our previous discussion, having a large value of $\alpha$ in BL-GU also tends to select continuous road segments on the long trajectories, which leads to a  different bike lane system in Figure \ref{sfig:gu110}. Interestingly, the selected bike lanes from  Figure \ref{sfig:gu110} are mostly aligned with the main roads of the city road network connecting different districts while the selected bike lanes from Figure \ref{sfig:ac2} are more along with the secondary roads.

\subsubsection{Coverage-Continuity Trade-Off.}  In both BL-AC and BL-GU, we are balancing the coverage objective versus the continuity objective, which are reflected in the utility functions. When increasing the value of $\lambda$ or $\alpha$, the bike lane planning model puts more weights on the continuity than the coverage objective. Table \ref{tab:tradeoff} presents the percentage changes of the coverage and continuity measures using BL-AC with $\lambda=0$ as the baseline. The coverage ratio is calculated as the percentage of trajectories (sets of road segments) that are covered by bike lanes. Indeed, we observe that both the mean number of connections per bike lane and mean size of continuous bike lanes grow at the expense of the coverage ratio. Notably, the BL-GU with $\alpha=1.02$ improves the continuity of bike lanes significantly while only lowering the coverage ratio slightly.

\begin{table}
	\centering
	\caption{Percentage change in coverage and continuity measures with varying $\lambda$ and $\alpha$.}
	\begin{tabular}{cccc}
		\toprule
		&  \multicolumn{1}{p{3.2cm}}{\centering \% change in the coverage ratio} &  \multicolumn{1}{p{5.5cm}}{\centering \% change in the mean \# of connections per bike lane} &  \multicolumn{1}{p{5.5cm}}{\centering \% change in the mean size of continuous bike lanes}  \\
		\midrule
		\multicolumn{4}{c}{BL-AC} \\
		$\lambda=2$     & -7.15\%  & 56.76\%  & 42.42\% \\
		$\lambda =10$    & -10.16\% & 64.86\%  & 54.55\% \\
		\multicolumn{4}{c}{BL-GU} \\
		$\alpha=1.02$  & -2.26\%  & 18.92\%  & 33.33\% \\
		$\alpha = 1.05$  & -20.46\% & 51.35\%  & 78.79\% \\
		\bottomrule
	\end{tabular}%
	\label{tab:tradeoff}%
\end{table}%

\subsubsection{Varying the Budget, $B$.} 
\begin{figure}
	\subfloat[Coverage ratio versus $B$ (km)\label{sfig:cover_B}]{
		\includegraphics[width=0.5\textwidth]{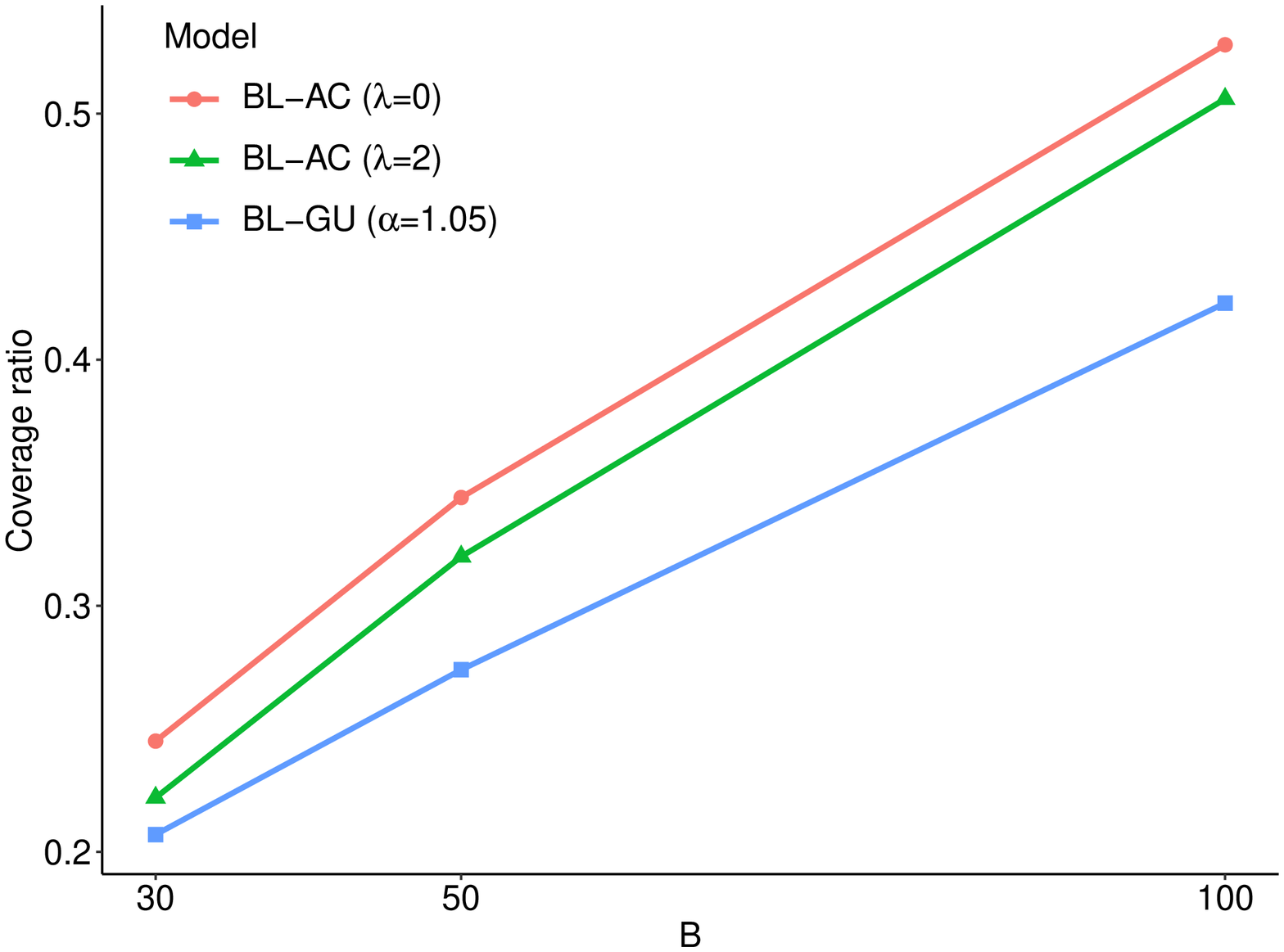}
	}\hfill
	\subfloat[Mean size of continuous bike lanes versus $B$ (km)\label{sfig:mean_size_B}]{
		\includegraphics[width=0.5\textwidth]{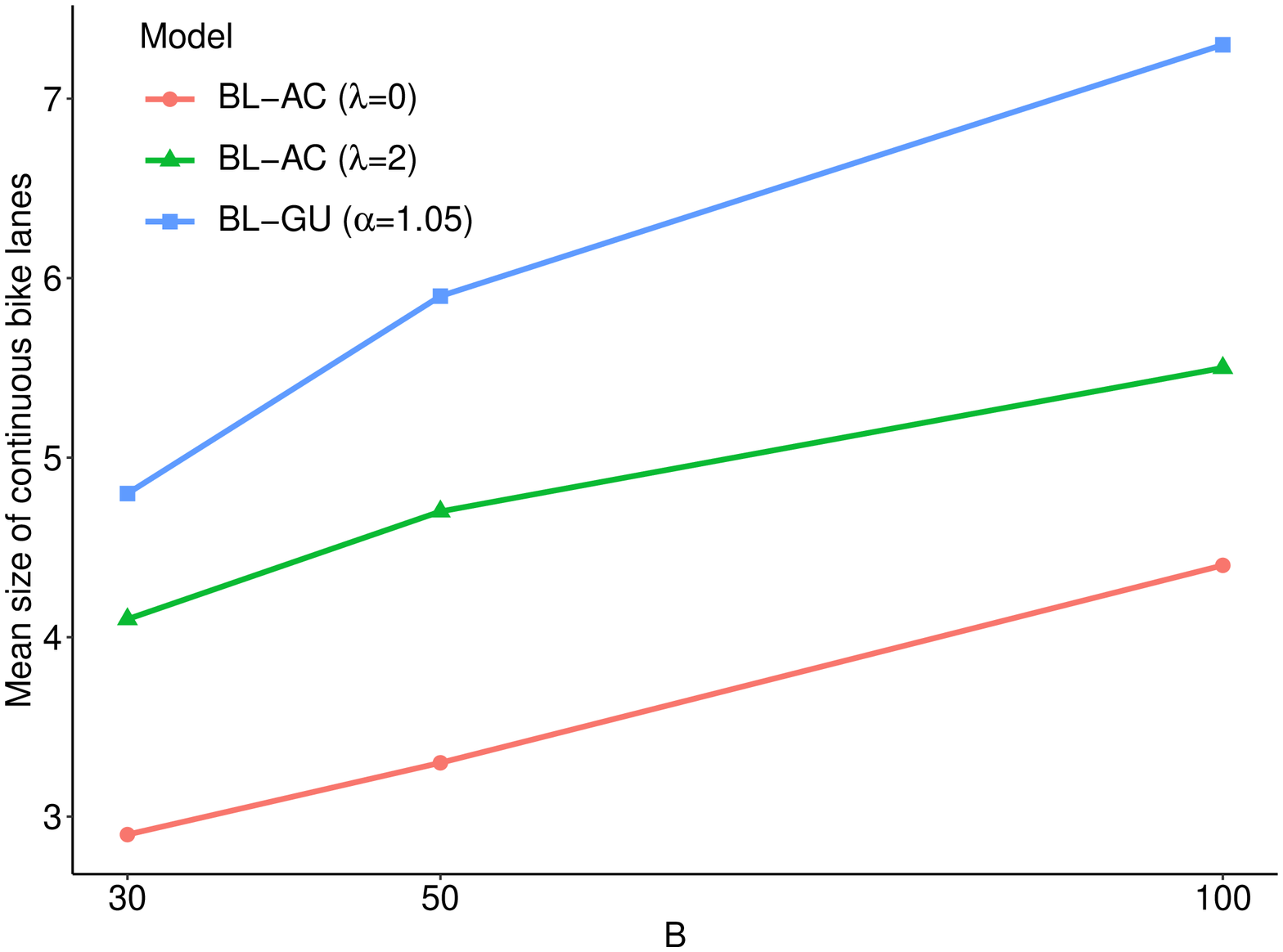}
	}\hfill
	\caption{The impact of $B$ on the coverage ratio and mean size of continuous bike lanes.}\label{fig:result_B}
\end{figure}

Increasing the budget in both BL-AC and BL-GU will improve coverage and continuity. Figure \ref{fig:result_B} shows how the coverage ratio and the mean size of continuous bike lanes evolve with the budget value. As $B$ grows, the two measures increase and the gap between different models remains non-negligible. This implies that even when a large budget is allowed, making a good bike lane planning decision is still critical. Since bike lane construction is often costly, the city government can weigh the benefits from building bike lanes versus the cost using the quantified measures proposed in this paper.

\subsubsection{Route choice model considerations.} \hl{The previous discussion ignores the interaction between bike lane planning and cyclists' route choices. Here we investigate how the planning result would change when cyclists update their route choices following the discussion in Section \ref{subsec:behav}.} To better illustrate and compare the result, we focus on a subarea of the city (Jida Residential District), which corresponds to the bottom right area on the map. We call the Google Map Directions API (\url{https://developers.google.com/maps/documentation/directions/start}) to generate the candidate routes for each origin-destination pair in the subarea. Hence the choice set $C_m$ includes both the observed route as well as the routes returned by Google.

\hl{We apply the MILP model developed in Subsection \ref{subsec:behav} and choose the linear approximation sample size to be 20. 
The utility function is assumed to take the form of $v_x(r) - \eta L_r $ for a route $r$, where $L_r$ is the riding distance (in km) of route $r$ (since travel time is an important factor in cyclists' choices, see, e.g., \citealt{khatri2016modeling, datner2019setting}). The parameter $\eta$ measures the disutility per km of riding distance, which can be estimated following the MNL model. Here we set $\eta = 20$ such that $v_x(r)$ is comparable with $\eta L_r$\footnote{It is hard to identify $v_x(r)$ from the current data set so we may revisit the estimation problem once we collect new data sets after Zhuhai city implements more bike lanes.}.} 

\begin{figure}[ht]
	\subfloat[Without route choices ($\alpha=1.02$)\label{sfig:respons_no1}]{
		\includegraphics[width=0.43\textwidth, trim={9.6cm 4cm 4cm 6cm}, clip, keepaspectratio]{graphs_2020/select_points_a102_B50000_n6666.pdf}
	}\hfill
	\subfloat[Route choice based model ($\alpha=1.02$)\label{sfig:respons_1}]{
		\includegraphics[width=0.43\textwidth, trim={9.6cm 4cm 4cm 6cm}, clip, keepaspectratio]{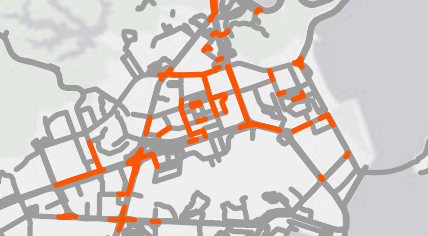}
	}\hfill
	\subfloat[Without route choices ($\alpha=1.05$)\label{sfig:respons_no2}]{
		\includegraphics[width=0.43\textwidth, trim={9.6cm 4cm 4cm 6cm}, clip, keepaspectratio]{graphs_2020/select_points_a105_B50000_n6666.pdf}
	}\hfill
	\subfloat[Route choice based model ($\alpha=1.05$)\label{sfig:respons_2}]{
		\includegraphics[width=0.43\textwidth, trim={9.6cm 4cm 4cm 6cm}, clip, keepaspectratio]{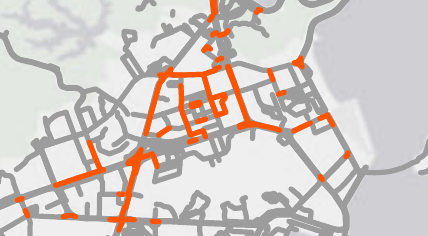}
	}\hfill
	\caption{\hl{Selected bike lanes from BL-GU with different responsive behaviors of cyclists ($B=50$ km)}}\label{fig:result_response}
\end{figure}

\hl{Figure \ref{fig:result_response} presents the bike lane planning results for BL-GU with route choices ($B=30$ and $\alpha=1.02, 1.05$). We observe that accounting for route choices has a significant impact on the resulting bike lane network. Specifically, after incorporating the route choice model, the optimal bike lane network tends to spread out to cover more routes other than the observed ones, which compromises continuity to some extent.} This highlights the importance of understanding cyclists' routing choices when bike lanes are present (including both coverage and continuity considerations). \hl{Policy makers can employ empirical approaches to understand the behaviors and integrate them with the analytical models proposed in this paper. } 

\section{Conclusion}\label{sec:con}
This paper develops a set of novel formulations and algorithms to the bike lane planning problem that is becoming increasingly important in smart city management. Unlike the previous work that mainly builds on surveys and heuristics, we present a modeling framework that directly utilizes the GPS bike trajectory data from emerging dock-less bike sharing systems. Our model formalizes the main objectives of bike lane planning in view of the cyclists' utility functions. Depending on the utility function structure, we propose efficient algorithms to solve the corresponding bike lane planning model. \hl{In addition, we develop a route choice based bike lane planning model that considers the interaction between bike lane network and cyclists' routing behaviors. Based on the bilevel program reformulation, we obtain a tractable MILP reformulation.} We demonstrate the effectiveness of the proposed models and algorithms on a large-scale real-world data set. We show how the topology of the bike lane network would change with varying choices of the utility functions and demonstrate the tension between coverage and continuity quantitatively. Last but not least, our results reveal the importance of considering cyclists' route choices to the bike lane planning, which is often ignored in the existing literature.

There are several promising future research directions. First, it would be interesting to consider the width of bike lanes as another decision dimension in the model. The width of the bike lanes can impact the interaction between car flows and bike flows, and eventually change the traffic equilibrium of the whole urban environment. Second, since cyclists' responsive behaviors are often hard to predict before any bike lane is deployed, the city planning agencies may dynamically construct bike lanes in the city, e.g., first build a few bike lanes to learn the behaviors and then add more bike lanes. Then the problem becomes a dynamic strategic planning model with behavior learning. Third, jointly designing the bike lanes with other bike facilities such as bike sharing stations and parking areas can be of interest to the municipal governments. 
\bibliographystyle{ormsv080}
\bibliography{bike_ref}

\newpage
\section*{Appendix}
\hl{In this appendix we provide the technical proofs of the results in the paper as well as additional modeling details (the MILP reformulation to BL-GU with route choices, and data decensoring steps).}
\begin{APPENDICES}
\section{Main proofs}
\begin{proof}{Proof of Lemma \ref{lem:1}}
	Let two sets of road segments $\mathcal{A}$ and $\mathcal{B}$ satisfy $\mathcal{A}\subseteq \mathcal{B}$. Consider adding a road segment $k\notin B$ to the two sets. In terms of the first part of the objective function, the change caused by adding $k$ is the same for $\mathcal{A}$ and $\mathcal{B}$ because of linearity. For the second part, as $\lambda\geq0$ and $d_{ij}\geq0$ for all $(i,j)\in N$, the value of the second part can only increase with the newly added $e$. Since $\mathcal{B}$ can only contain more road segments than $\mathcal{A}$, there are potentially more neighbors of $k$ that are included in $\mathcal{B}$. As a result, the increase of objective value caused by adding $e$ is greater for $\mathcal{B}$ than $\mathcal{A}$. \Halmos
\end{proof} 

\begin{proof}{Proof of Proposition \ref{prop:1}} Directly from Lemma \ref{lem:1}, \ref{obj:form1} is  essentially a supermodular knapsack problem, which admits a polynomial-time solvable Lagrangian dual, as shown in \hl{Theorem 16} of \cite{gallo1989supermodular}.\Halmos
\end{proof}

\begin{proof}{Proof of Theorem \ref{th:1}} Consider two bike lane construction plans, i.e. sets of road segments $\mathcal{A}$ and $\mathcal{B}$ to build bike lanes satisfying $\mathcal{A}\subseteq \mathcal{B}$, and a road segment $i\notin \mathcal{B}$. We prove the supermularity of $v_x(r)$ by conditioning on $i$: 1) If $i\notin r$, then building a bike lane on $i$ does not impact the value of $v_x(r)$, so $v_{\mathcal{A}\cup \{i\}}(r) = v_A(r)$ and $v_{\mathcal{B}\cup \{i\}}(r) = v_B(r)$. 2) If $i\in r$, let $i-1$ and $i+1$ denote the road segment visited before and after $i$ on the trajectory $r$, respectively (when $i$ is at the head or the tail of the trajectory, either $i-1$ or $i+1$ is empty and our analysis can be easily extended). We further consider the following three scenarios:
\begin{itemize}
	\item If \hl{neither $i-1$ nor $i+1$ belongs to $\mathcal{A}$, then $v_{\mathcal{A}\cup \{i\}}(r) - v_{\mathcal{A}}(r) = f(1)$. Next, if either $i-1$ or $i+1$ belongs to $\mathcal{B}$, we have $v_{\mathcal{B}\cup \{i\}}(r) - v_{\mathcal{B}}(r) = f(|s|+1) - f(|s|)$ for some $s\in S_{\mathcal{B}}(r)$ (note that $s$ must include either $i-1$ or $i+1$, thus $|s|\geq 1$). By the assumption that $f(\cdot)$ is increasing convex, we have $f(|s|+1) - f(|s|) \geq f(1)$. If both $i-1$ and $i+1$ belong to $\mathcal{B}$, then $v_{\mathcal{B}\cup \{i\}}(r) - v_{\mathcal{B}}(r) = f(|s_{-1}| + |s_{+1}| + 1) - f(|s_{-1}|) - f(|s_{+1}|)$, where $s_{-1},s_{+1} \in S_{\mathcal{B}}(r)$ and $i-1\in s_{-1}$, $i+1\in s_{+1}$. Because $f(\cdot)$ is increasing convex, $f(|s_{-1}| + |s_{+1}| + 1) - f(|s_{-1}|) - f(|s_{+1}|) \geq f(|s_{-1}| + |s_{+1}| + 1) - f(|s_{-1}| + |s_{+1}|) \geq f(1)$.} Otherwise, when neither $i-1$ nor $i+1$ belongs to $\mathcal{B}$, $v_{\mathcal{B}\cup \{i\}}(r) - v_{\mathcal{B}}(r) = f(1)= v_{\mathcal{A}\cup \{i\}}(r) - v_{\mathcal{A}}(r)$. Hence $v_{\mathcal{B}\cup \{i\}}(r) - v_{\mathcal{B}}(r)\geq v_{\mathcal{A}\cup \{i\}}(r) - v_{\mathcal{A}}(r)$. 
	
	\item \hl{If either $i-1$ or $i+1$ belongs to $\mathcal{A}$, then  $v_{\mathcal{A}\cup \{i\}}(r) - v_{\mathcal{A}}(r) = f(|s|+1) - f(|s|)$ for some $s\in S_{\mathcal{A}}(r)$. Similarly, for $\mathcal{B}$, as $\mathcal{A}\subseteq \mathcal{B}$ we have: i) when  either $i-1$ or $i+1$ belongs to $\mathcal{B}$, $v_{\mathcal{B}\cup \{i\}}(r) - v_{\mathcal{B}}(r) = f(|s'|+1) - f(|s'|)$ for some $s'\in S_{\mathcal{B}}(r)$; ii) when  both $i-1$ and $i+1$ belong to $B$, $v_{\mathcal{B}\cup \{i\}}(r) - v_{\mathcal{B}}(r) = f(|s_{-1}'| + |s_{+1}'| + 1) - f(|s_{-1}'|) - f(|s_{+1}'|)$, where $s_{-1}',s_{+1}' \in S_{\mathcal{B}}(r)$ and $i-1\in s_{-1}'$, $i+1\in s_{+1}'$. Because $\mathcal{A}\subseteq \mathcal{B}$, we can verify that $|s'| \geq |s|$ and $|s_{-1}'| + |s_{+1}'| > |s|$.  As $f(\cdot)$ is increasing convex, $v_{\mathcal{B}\cup \{i\}}(r) - v_{\mathcal{B}}(r)\geq v_{\mathcal{A}\cup \{i\}}(r) - v_{\mathcal{A}}(r)$ . }
	
	\item \hl{If both $i-1$ and $i+1$ belong to $A$, $v_{\mathcal{A}\cup \{i\}}(r) - v_{\mathcal{A}}(r) =f(|s_{-1}| + |s_{+1}| + 1) - f(|s_{-1}|) - f(|s_{+1}|)$, where $s_{-1},s_{+1} \in S_{\mathcal{A}}(r)$ and $i-1\in s_{-1}$, $i+1\in s_{+1}$. Similarly, both $i-1$ and $i+1$ must belong to $B$ as well, hence $v_{\mathcal{B}\cup \{i\}}(r) - v_{\mathcal{B}}(r) = f(|s_{-1}'| + |s_{+1}'| + 1) - f(|s_{-1}'|) - f(|s_{+1}'|)$, where $s_{-1}',s_{+1}' \in S_{\mathcal{B}}(r)$ and $i-1\in s_{-1}'$, $i+1\in s_{+1}'$. Because $\mathcal{A}\subseteq \mathcal{B}$, $|s_{-1}'| \geq |s_{-1}|$ and  $|s_{+1}'| \geq |s_{+1}|$. It follows that $|s_{-1}'| + |s_{+1}'| \geq  |s_{-1}| + |s_{+1}|$ and $v_{\mathcal{B}\cup \{i\}}(r) - v_{\mathcal{B}}(r) \geq v_{\mathcal{A}\cup \{i\}}(r) - v_{\mathcal{A}}(r)$.}
\end{itemize} 
In all the cases we have shown that $v_{\mathcal{B}\cup \{i\}}(r) - v_{\mathcal{B}}(r)\geq v_{\mathcal{A}\cup \{i\}}(r) - v_{\mathcal{A}}(r)$ so $v_x(r)$ is indeed supermodular. \Halmos
\end{proof}

\begin{proof}{Proof of Corollary \ref{cor:lagdual}} Based on Theorem \ref{th:1}, the utility function $v_x(r)$ is supermodular and maximization of this utility function over a budget constraint is a supermodular knapsack problem, which has a polynomial-time solvable Lagrangian dual \hl{(see Theorem 16 from \citealt{gallo1989supermodular})}.\Halmos
\end{proof}

\begin{proof}{Proof of Proposition \ref{prop:genmilp}} Following the reformulation of $v_x(r)$ as a polynomial function (\ref{eqn:generalu}), we can further apply standard linearization techniques to get an MILP formulation, as shown in BL-GU-MILP.  \Halmos
\end{proof}
	
\begin{proof}{Proof of Proposition \ref{prop:milptu}} When $f(\cdot)$ is an increasing convex function, the coefficients $\beta_l$ in BL-GU-MILP are all positive: $\beta_l =  f(|l|) - 2f(|l|-1) + f(|l|-2) = f(|l|) - f(|l|-1) - (f(|l|-1) - f(|l|-2)) \geq 0$. Then we can remove constraints (\ref{const:gen1}) from the formulation. The remaining constraints are constraints (\ref{const:gen2}) and constraints (\ref{const:gen4}) in addition to the budget constraints. Our goal is to prove that the constraint matrix consisting of these two types of constraints is totally unimodular. 
Note that constraints (\ref{const:gen4}) correspond to an identity matrix and do not affect the totally unimodularity property. Therefore we focus on the matrix of constraints (\ref{const:gen2}), which we denote by $P$. With or without using the reduction techniques, each row of $P$ only includes one $1$ and one $-1$. Then we prove that $P^T$ is totally unimodular because it satisfies: 1) Each entry of $P^T$ \hl{must be -1, 0, or +1, i.e., $P_{ij}^T\in\{-1,0,+1\}$}; 2) Each column contains at most two non-zero entries; 3) There exists a partition $(M_1, M_2)$ of the set $M$ of the set of rows of $P^T$ such that each column $j$ containing two non-zero entries satisfies $\sum_{i\in M_1}P_{ij}^T - \sum_{i\in M_2}P_{ij}^T=0$. It is obvious that conditions 1) and 2) hold for $P^T$, and the third condition is satisfied by using the partition $(P^T, \emptyset)$. It follows that $P$ is totally unimodular because the transpose of a totally unimodular matrix is also totally unimodular. As a result, the Lagrangian relaxation of BL-GU-MILP with relaxed budget constraint can be solved as an LP. \Halmos
\end{proof}

\begin{proof}{Proof of Proposition \ref{prop:approx}}
\hl{
Note that the approximation error only comes from the piecewise linear approximation of function $p\ln p$ in the lower-level problem (LL). Let $\mathbf{p}$ be the solution to (LL) and $\mathbf{p}'$ be the solution to (LL-Lin) (given decision $x$). And we use $h(\cdot)$ and $h'(\cdot)$ to denote the objective function of (LL) and (LL-Lin), respectively. It follows that
\[ \nabla h(\mathbf{p}) = (\ln p_{mr} + v_x(r) - \bar{v}(r))_{mr}.   \]
It can be easily verified that $\nabla h(\mathbf{p})$ is strongly monotone with modulus $1$ for $0< \mathbf{p}\leq 1$ (the Jacobian matrix is positive definite diagonal, and the smallest possible eigenvalue is $1$, see also \citealt{dan2019competitive}). From the definitions of $\mathbf{p}$ and $\mathbf{p}'$, we have
\begin{align*}
 &\langle\nabla h(\mathbf{p}),  \mathbf{p}' - \mathbf{p}\rangle \geq 0,\ \langle\nabla h'(\mathbf{p}'),  \mathbf{p} - \mathbf{p}'\rangle \geq 0\\
 \Rightarrow\ & \langle \nabla h'(\mathbf{p}') - \nabla h(\mathbf{p}),  \mathbf{p} - \mathbf{p}'\rangle \geq 0. 
\end{align*}
By the properties of $\nabla h(\cdot)$, it holds that $\langle\nabla h(\mathbf{p}) - \nabla h(\mathbf{p}'), \mathbf{p} - \mathbf{p}' \rangle \geq ||\mathbf{p} - \mathbf{p}'||^2$ (from the Jacobian matrix), hence 
\begin{equation} \label{eqn:append1}
 \langle \nabla h'(\mathbf{p}') - \nabla h(\mathbf{p}'),  \mathbf{p} - \mathbf{p}'\rangle \geq  ||\mathbf{p} - \mathbf{p}'||^2.
\end{equation}
Furthermore, the incurred difference in the objective function of the upper-level problem (\ref{eqn:obj_choice}) is
\begin{align}
\sum_{m\in M} \sum_{r\in C_m} D_m\left|(p_{mr} - p'_{mr}) (v_x(r) - \bar{v}(r)) \right| & \leq D \cdot V\sum_{m\in M} \sum_{r\in C_m} \left|p_{mr} - p'_{mr}\right| \nonumber\\
& \leq D \cdot V\sqrt{|M|\cdot|C^*|}\ || \mathbf{p} - \mathbf{p}' || \nonumber \\
& \leq D \cdot V\sqrt{|M|\cdot|C^*|}\ || \nabla h'(\mathbf{p}') - \nabla h(\mathbf{p}') ||, \label{eqn:prop42}
\end{align}
where $D = \max_m D_m$, $V = \max_{x, r} |v_x(r) - \bar{v}(r)|$, and $C^* = \arg\max_m |C_m|$. The second inequality follows directly from the Cauchy–Schwarz inequality, and the third inequality is derived by applying the Cauchy–Schwarz inequality to (\ref{eqn:append1}). Next, we bound the gap between $\nabla h'(\mathbf{p}')$ and $\nabla h(\mathbf{p}')$.}

\hl{
By the construction of $h'(\cdot)$, each component of $\nabla h'(\mathbf{p}')$, $\nabla h'(\mathbf{p}')_{mr}$, is a piecewise constant approximation of $\log(p_{mr})$. Let the $K$ sampling points $\{p^1, p^2, \dots, p^K\}$ satisfy: 1) the samples start from $p_{\min}$ to $1$; and 2) the sample points are chosen such that the constant approximations are vertically equidistant. Then we have
\[ \left|\nabla h'(\mathbf{p}')_{mr} - \nabla h(\mathbf{p}')_{mr} \right| \leq \frac{ \log\frac{1}{p_{\min}}}{K}.  \]
Therefore, 
\begin{align}
|| \nabla h'(\mathbf{p}') - \nabla h(\mathbf{p}') || & = \sqrt{\sum_{m\in M} \sum_{r\in C_m} \left|\nabla h'(\mathbf{p}')_{mr} - \nabla h(\mathbf{p}')_{mr} \right|^2 } \leq \frac{ \log\frac{1}{p_{\min}} \sqrt{|M|\cdot|C^*|} }{K}. \label{eqn:prop43}
\end{align}
Combining inequalities (\ref{eqn:prop42}) and (\ref{eqn:prop43}) leads to
\[ \sum_{m\in M} \sum_{r\in C_m} D_m \left|(p_{mr} - p'_{mr}) (v_x(r) - \bar{v}(r)) \right| \leq \frac{  D \cdot V\cdot |M|\cdot|C^*|\log\frac{1}{p_{\min}}}{K}.  \] 
Thus the approximation error of the upper-level objective function (\ref{eqn:obj_choice}) is $O(\frac{1}{K})$. \Halmos}
\end{proof}

\section{The validity about the calculation of the coefficients $\beta$ in general utility functions} \label{appen:beta}
\hl{
For any trajectory $r=\{i^1, \dots, i^n\}$ ($n\in \mathbb{N}^+$), the chosen $\beta$'s  will need to satisfy
\begin{equation}\label{eqn:beta}
f(|r|)  =  \beta_r +  \sum_{j=1}^2 \beta_{i^j, \dots, i^{n+j-2}}  + \sum_{j=1}^3 \beta_{i^j, \dots, i^{n+j-3}} + \dots + \sum_{j=1}^n \beta_{i^j}.
\end{equation}
We will prove this by induction. It is easy to verify that the calculation presented in Subsection (\ref{sec:gu}) satisfy the above equations for trajectories with cardinality less or equal to three. Now we assume equations (\ref{eqn:beta}) are satisfied with $\beta_l = f(|l|) - f(|\{i^j, i^{j+1}, \dots, i^{j+k-1}\}|) - f(|\{i^{j+1}, i^{j+2}, \dots, i^{j+k}\}|) + f(|\{i^{j+1}, \dots, i^{j+k-1}\}|)$ for trajectories with cardinality less or equal to $n-1$. Then for a trajectory $r = \{i^1, \dots, i^n\}$, we have
\begin{align}
& \beta_r + \sum_{j=1}^2 \beta_{i^j, \dots, i^{n+j-2}}  + \sum_{j=1}^3 \beta_{i^j, \dots, i^{n+j-3}} + \dots + \sum_{j=1}^n \beta_{i^j} \nonumber \\
= & \beta_r + \left[ \beta_{i^2, \dots, i^{n}} + \sum_{j=2}^3 \beta_{i^j, \dots, i^{n+j-3}} +\dots +  \sum_{j=2}^n \beta_{i^j}    \right] + \beta_{i^1, \dots, i^{n-1}} + \beta_{i^1, \dots, i^{n-2}} + \dots + \beta_{i^1} \nonumber \\
= &  \beta_r + f(|\{i^2, \dots, i^{n}\}|) + \beta_{i^1, \dots, i^{n-1}} + \beta_{i^1, \dots, i^{n-2}} +\dots + \beta_{i^1} , \label{eqn:betas}
\end{align}
where the last equality follows by our inductive assumption. Furthermore, note that
\begin{align}
\beta_{i^1} +  \beta_{i^1,i^2} +  \dots + \beta_{i^1, \dots, i^{n-2}}  + \beta_{i^1, \dots, i^{n-1}}  = & f(|\{i^1,i^2\}|) - f(|\{i^2\}|) + \beta_{i^1,i^2,i^3} + \dots + \beta_{i^1, \dots, i^{n-1}} \nonumber \\
= & f(|\{i^1,i^2, i^3\}|) - f(|\{i^2, i^3\}|) + \dots + \beta_{i^1, \dots, i^{n-1}}\ \nonumber \
& \vdots  \nonumber \\
= & f(|\{i^1, \dots, i^{n-1}\}) - f(|\{i^2, \dots, i^{n-1}\}),\label{eqn:betas2}
\end{align}
where the equalities hold by our choices of $\beta$'s. Combining the equations (\ref{eqn:betas}) and (\ref{eqn:betas2}) gives
\[  \beta_r + \sum_{j=1}^2 \beta_{i^j, \dots, i^{n+j-2}}  + \sum_{j=1}^3 \beta_{i^j, \dots, i^{n+j-3}} + \dots + \sum_{j=1}^n \beta_{i^j} = \beta_r + f(|\{i^2, \dots, i^{n}\}|) + f(|\{i^1, \dots, i^{n-1}\}) - f(|\{i^2, \dots, i^{n-1}\}) = f(|r|). \]
Hence equations (\ref{eqn:beta}) hold for trajectories with cardinality equals to $n$. \Halmos
}
\section{The detailed MILP formulation of the bike lane planning model with route choices} \label{appen:lin}
\hl{
The optimality equations for (LL-Lin) can be written as
\begin{align}
& (\ref{constr:lin1})-(\ref{constr:lin3}) \nonumber \\
& \gamma_m - \sum_{k=1}^K \varrho_{mrk} \leq \bar{v}(r) - v_x(r), \ \forall m\in M, r \in C_m, \label{constr:lindual1}\\
& \sum_{k=1}^K \varrho_{mrk} = 1, \ \forall m\in M, r \in C_m, \label{constr:lindual2}\\
& \varrho_{mrk} \geq 0,  \ \forall m\in M, r \in C_m, \forall k= 1,\dots K \label{constr:lindual3}\\
& \sum_{m\in M}\sum_{r\in C_m}\big[ \omega_{mr} + p_{mr}\left(\bar{v}(r) - v_x(r) \right)  \big] = \sum_{m\in M} \gamma_m - \sum_{m\in M}\sum_{r\in C_m}\sum_{k=1}^K p^k \varrho_{mrk} \label{constr:lindual4}
\end{align}
where $\gamma_m$, $\varrho_{mrk}$ are the dual variables of constraints (\ref{constr:lin1}) and (\ref{constr:lin2}), respectively. Constraints (\ref{constr:lindual1})-(\ref{constr:lindual3}) are the dual feasible constraints. Constraint (\ref{constr:lindual4}) enforces the equivalence between the primal objective value and the dual objective value, which replaces the complementary slackness constraints. This replacement leads to better empirical computational performance as it avoids the use of big $M$ and additional binary variables, which would otherwise be required in linearizing the complementary slackness constraints.
}
\hl{
Then we introduce $\varphi_{mr} = p_{mr} v_x(r)$ with $\zeta_{mrl}$ and
\begin{align}
& \varphi_{mr} =  \sum_{l\in L(r)} \beta_l \zeta_{mrl},\ \forall  m\in M, r \in C_m \\
&\zeta_{mrl} \leq p_{mr},\quad \forall m\in M, r \in C_m, l\in L(r),\\
&\zeta_{mrl} \leq y_l,\quad \forall m\in M, r \in C_m, l\in L(r).
\end{align}
Because the objective is maximizing cyclists' utility, the above constraints will make $\zeta_{mrl} = p_{mr}y_l$. Therefore, we obtain an MILP formulation for the bike lane planning model with route choices. }
\section{Temporal distribution of trajectory Data} \label{appen:temp}
Figure \ref{sfig:hrs} depicts the distribution of bike trajectories across different hours of a day. It can be observed that there are two demand peaks: one occurs in the morning (7:00 to 9:00 am) and the other occurs in the evening (5:00 to 8:00 pm). These two peaks correspond to the commute rush hours. We also note that the bike trip demand falls gradually after the evening peak and still remains substantial until midnight, which can be attributed to people who engage in leisure activities after work. Figure \ref{sfig:weekday} shows the distribution of bike trajectories across different days of a week. We observe that the bike trip demand is almost stable throughout the week while there are two small peaks on Mondays and Saturdays.

\begin{figure}[htbp]
	\subfloat[Hour distribution\label{sfig:hrs}]{
		\includegraphics[width=0.5\textwidth,keepaspectratio]{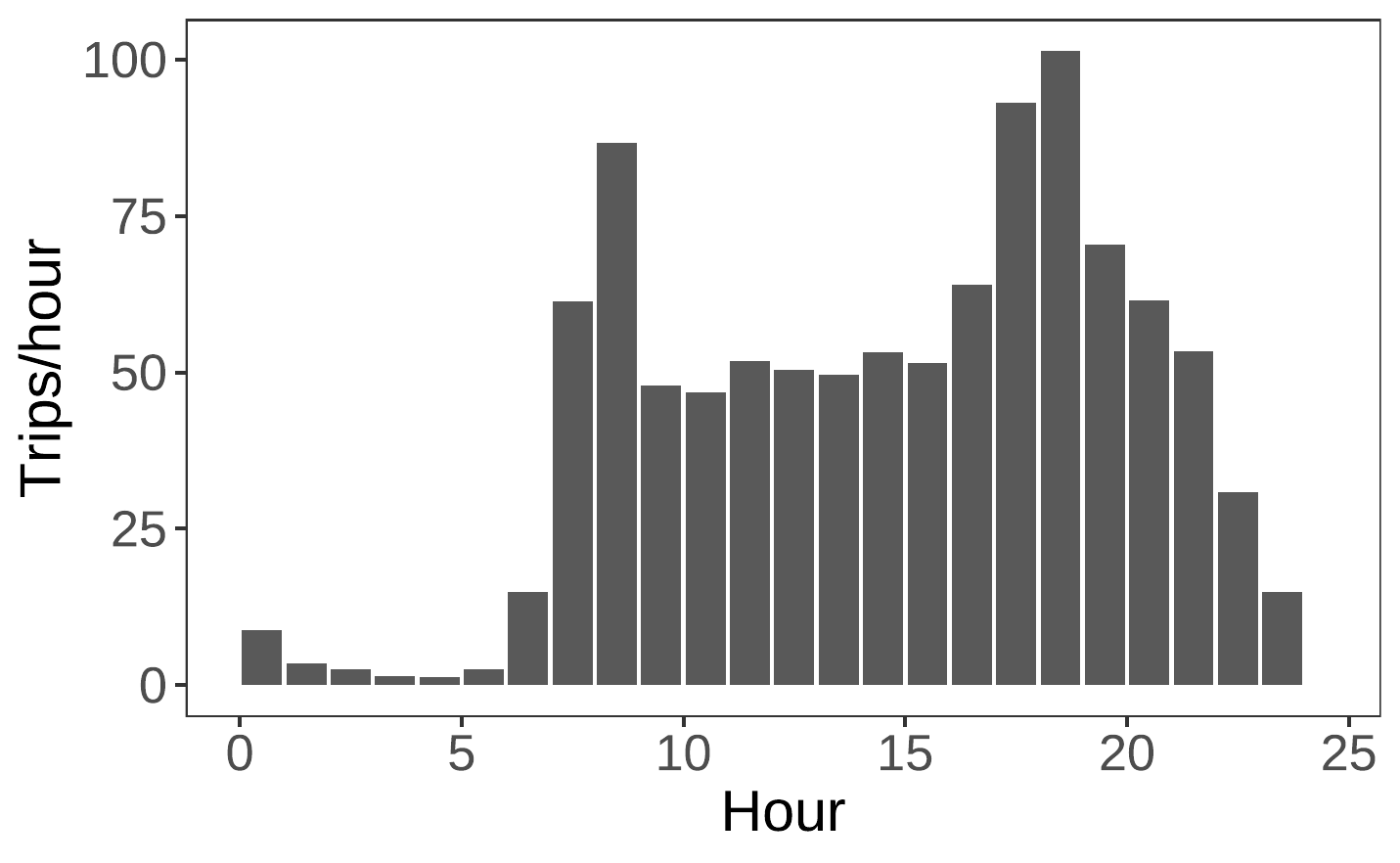}
	}\hfill
	\subfloat[Weekday distribution\label{sfig:weekday}]{
		\includegraphics[width=0.5\textwidth,keepaspectratio]{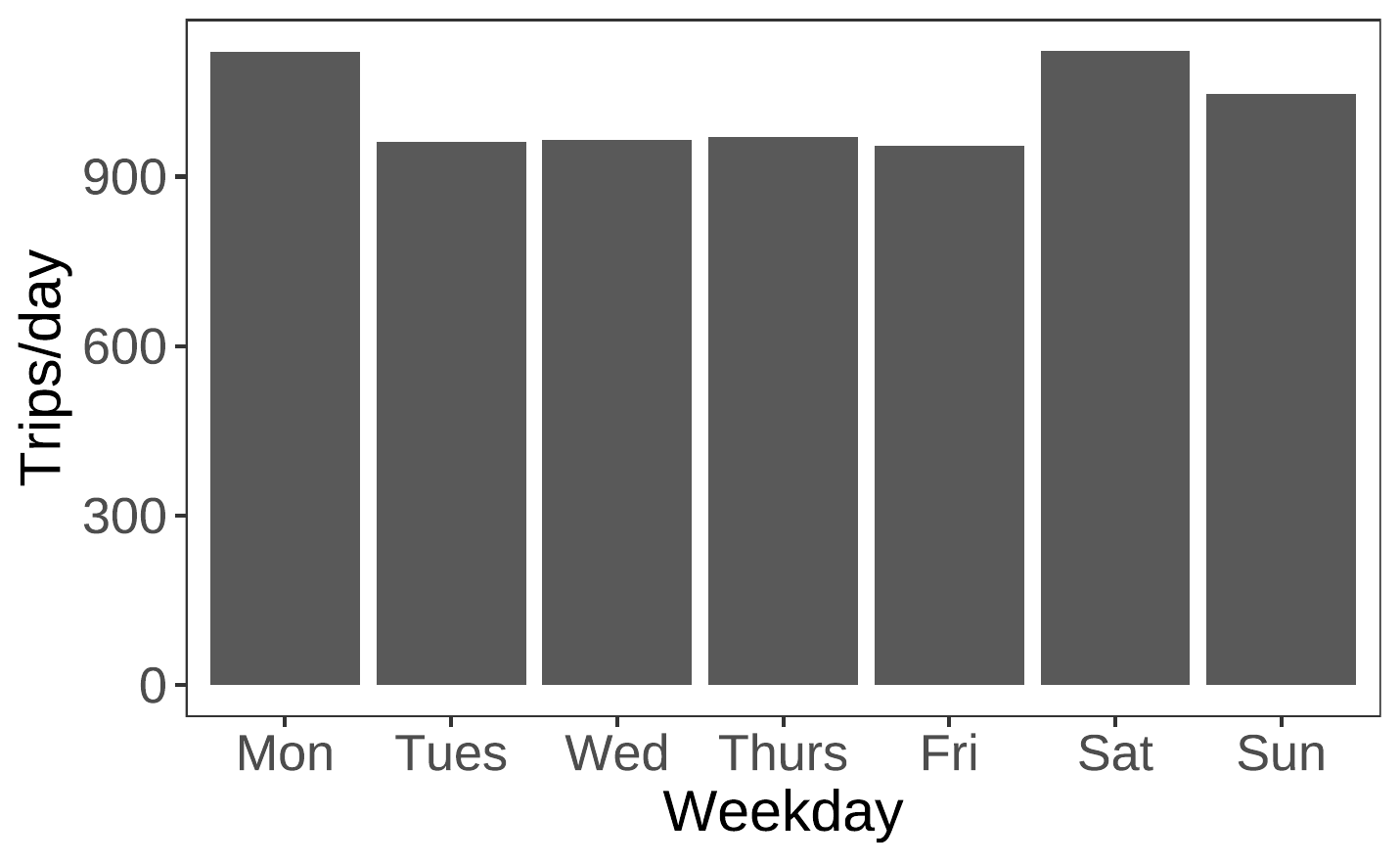}
	}
	\caption{Bike trajectory temporal distribution.}\label{fig:time}
\end{figure}

\section{The detailed decensoring procedure for bike trajectory data} \label{appen:decen}
\hl{To minimize the seasonality effect, we pick a two-week period with stable and high ride demand to perform decensoring operations. Because there are no docked stations and riders can find available bikes in their neighborhood with smart phones, we split the urban area into small neighborhoods using the geohash script, which is a widely used geocode system\footnote{For more information about the geohash system, see \url{https://www.movable-type.co.uk/scripts/geohash.html}. It is used by mobility service providers such as Lyft: \url{https://eng.lyft.com/matchmaking-in-lyft-line-691a1a32a008}}. Each neighborhood is about 153 meters by 153 meters according to geohash of length 7. Given a day, we track the evolution of available bikes (e.g., the stock level) every 10 minutes in each neighborhood. Then we identify the stock-out events from the stock evolution. A trajectory originating from neighborhood $i$ in period $j$ would be assigned a weight equals to $1/(\text{\# of days with no stock-outs in period j from neighborhood $i$})$. After aggregation the weighted trajectory data  essentially represents the average ride demand conditional on there are no stock-outs, which shares a similar rationale of \cite{o2015data}. }
\end{APPENDICES}
%
%
%






\end{document}